\documentclass[a4paper,fleqn]{cas-dc}

\usepackage[numbers]{natbib}
\usepackage{tabularx}
\usepackage{subcaption}

\def\tsc#1{\csdef{#1}{\textsc{\lowercase{#1}}\xspace}}
\tsc{WGM}
\tsc{QE}
\tsc{EP}
\tsc{PMS}
\tsc{BEC}
\tsc{DE}

\begin{document}
\let\WriteBookmarks\relax
\def\floatpagepagefraction{1}
\def\textpagefraction{.001}
\shorttitle{Leveraging social media news}
\shortauthors{J.K. Krishnan et~al.}

\title [mode = title]{$\text{VG}^2$GT: Voxel-Gaussian Splatting Visual Geometry Grounded Transformer}                      



\author[1]{Yibin Zhao}[type=editor,
                      orcid=0009-0006-2025-5555
                        ]
\ead{Y20230063@mail.ecust.edu.cn}
\credit{Data curation, Writing, Methodology, Experiments}

\author[2]{Yihan Pan}[type=editor,
                      orcid=0009-0005-5095-4739
                        ]
\credit{Validation, Methodology}
\ead{pyh1614273862@163.com}

\author[1]{Jun Nan}[type=editor,
                        ]
\ead{Y20240039@mail.ecust.edu.cn}
\credit{Investigation, Visualization}

\author[1]{WenLi Yang}[type=editor,
                        ]
\ead{18831321205@163.com}
\credit{Software, Investigation}

\author[3]{Liwei Chen}[type=editor,
                      orcid=0000-0001-5938-983
                        ]
\ead{18916309291@163.com}
\credit{Investigation, Methodology}

\author[1]{Jianjun Yi}[type=editor,
                        ]
\cormark[1]
\ead{jjyi@ecust.edu.cn}
\credit{Funding acquisition, Supervision, Validation}


\affiliation[1]{organization={East China University of Science and Technology},
                addressline={130 Meilong Road}, 
                city={Shanghai},
                postcode={200237}, 
                country={China}}
\affiliation[2]{organization={Shanghai Open University},
                addressline={288 Guoshun Road}, 
                postcode={200433}, 
                city={Shanghai},
                country={China}}
\affiliation[3]{organization={Shanghai Xiaoyuan Innovation Center},
                addressline={599 Xingmei Road}, 
                postcode={200237}, 
                city={Shanghai},
                country={China}}

\cortext[cor1]{Corresponding author}


\begin{abstract}
Gaussian Splatting has shown strong potential for 3D reconstruction and novel view synthesis. However, most existing methods rely on accurate camera parameters and per-scene optimization. Feed-forward alternatives avoid this optimization, but pixel-aligned Gaussian primitives often lead to artifacts and non-uniform scene representations.
We propose $\text{VG}^2$GT, a Voxel-Gaussian Splatting Visual Geometry-Grounded Transformer. $\text{VG}^2$GT builds on a frozen pretrained visual foundation model (VFM) and introduces a multi-scale differentiable voxel module to improve geometric reasoning. The module enhances patch tokens at the coarse level and splits voxel features at the fine level to regress Gaussian primitive parameters. During training, stochastic solid volume rendering supervises scene geometry, enabling accurate Gaussian scene reconstruction while keeping the VFM fully frozen. This design allows $\text{VG}^2$GT to be plugged into patch-feature-based VFMs and substantially reduces training cost.
$\text{VG}^2$GT achieves state-of-the-art performance on the widely used DTU, Replica, TAT, and ScanNet datasets.
\end{abstract}




\maketitle

\begin{figure*}[pos=t]
\centering
\includegraphics[width=0.95\textwidth]{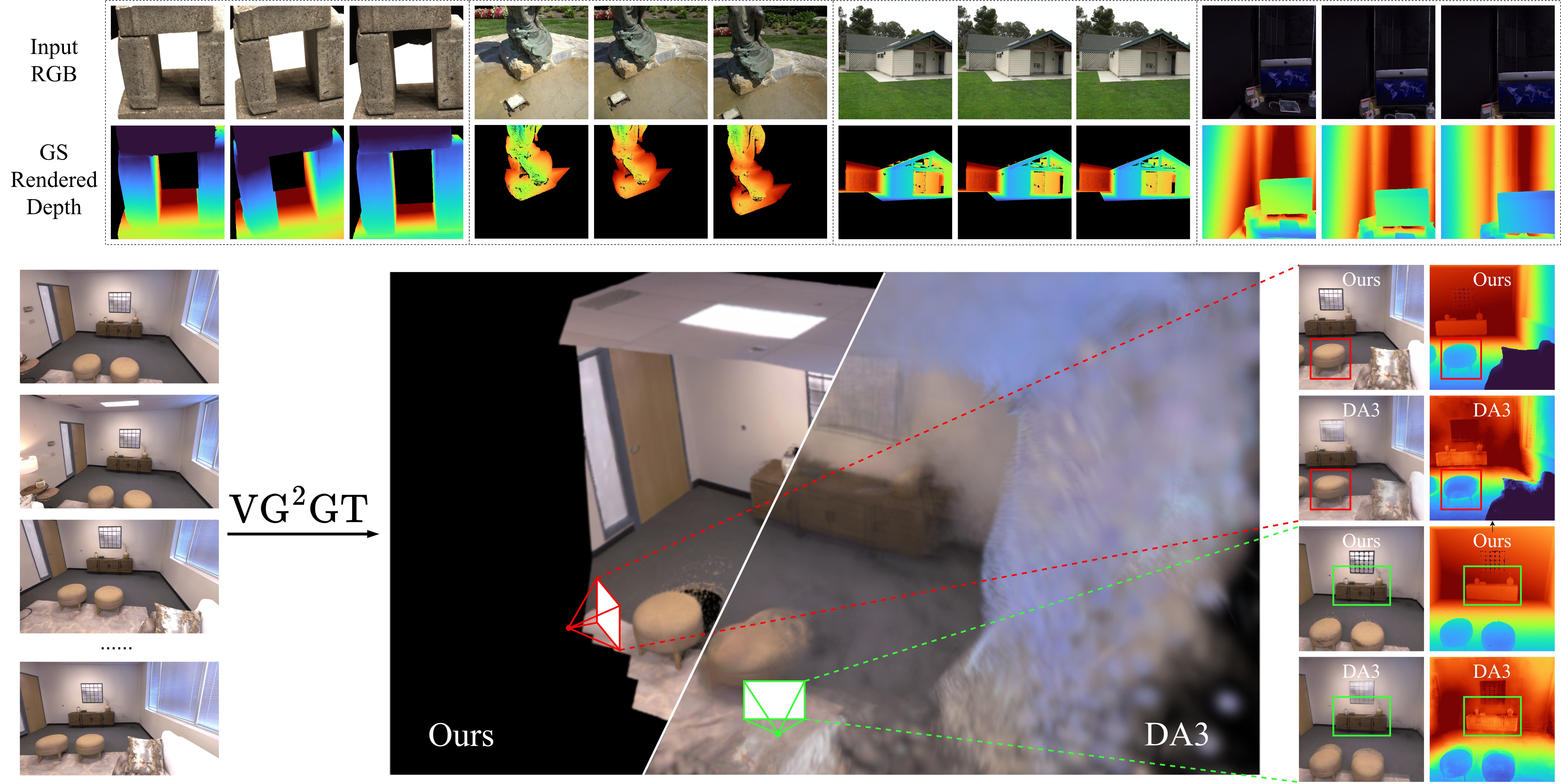}
\caption{\textbf{$\text{VG}^2$GT}, a non-pixel-aligned Gaussian Splatting \cite{kerbl3DGS} feed-forward model for fast 3D reconstruction and novel view synthesis. $\text{VG}^2$GT directly regresses geometrically accurate and spatially uniform Gaussian scenes from an arbitrary number of unposed and uncalibrated images within seconds.}
\label{fig_toutu}
\end{figure*}

\section{Introduction}
Recent visual foundation models \cite{wang2025vggt,depthanything3,dust3r_cvpr24} have shown strong potential for reconstructing 3D scenes from multi-view 2D images. By consolidating multi-stage reconstruction pipelines into a unified large-scale visual Transformer, these models improve efficiency and reduce error accumulation from hand-crafted intermediate steps.

Most visual foundation models represent reconstructed scenes as depth maps or point clouds, which provide strong geometric priors. However, point-cloud-based representations are less effective at preserving camera-level image details and view-dependent appearance. Radiance-field-based representations, including NeRF \cite{mildenhall2020nerf} and 3D Gaussian Splatting \cite{kerbl3DGS}, have therefore become powerful alternatives for novel view synthesis (NVS) and 3D reconstruction. These methods usually require known camera intrinsics and extrinsics. In practice, they often depend on structure-from-motion (SfM) to estimate camera parameters \cite{colmapsfm}, followed by per-scene optimization to recover the final scene representation. This pipeline propagates SfM errors into reconstruction and makes radiance-field-based methods computationally expensive.

Feed-forward 3D Gaussian Splatting (3DGS) reconstruction methods \cite{chen2024mvsplat,xu2024depthsplat,wang2025volsplat} offer a promising route to efficient and high-fidelity scene reconstruction. By avoiding per-scene optimization, they directly regress Gaussian scene representations from multi-view images and camera parameters. Nevertheless, these methods still require camera poses to construct multi-view cost volumes for depth estimation. This requirement limits their applicability to unconstrained real-world inputs and keeps them vulnerable to errors introduced by SfM.

The progress of visual foundation models has further enabled pose-free feed-forward 3DGS reconstruction \cite{depthanything3,jiang2025anysplat,ye2025yonosplatneedmodelfeedforward}. However, existing methods commonly use pixel-aligned Gaussian primitives, which impose a one-to-one correspondence between 2D pixels and 3D points. In real multi-view observations, the same 3D point can be visible from multiple pixels and viewpoints. Aggregating these observations is essential for building compact, coherent, and geometrically consistent scene representations. Pixel-aligned formulations instead tend to produce uneven Gaussian distributions and cross-view inconsistencies \cite{li2026tokensplat}, which limit reconstruction quality and multi-view consistency.

Here we propose $\text{VG}^2$GT, a feed-forward non-pixel-aligned 3DGS reconstruction framework. Given an arbitrary number of pose-free images, $\text{VG}^2$GT directly regresses geometrically accurate Gaussian scenes and camera parameters. To reduce the uneven distribution of Gaussian primitives, we introduce a multi-scale differentiable voxel module that aggregates point-cloud features into structured voxel features. At the coarse scale, PointTransformer \cite{wu2024ptv3} and trilinear interpolation compute complementary geometric features, which are injected into patch tokens to enhance global scene understanding. At the fine scale, a DPT head extracts global point-cloud features and voxelizes them into a structured representation. The resulting voxel features are refined by PointMLP and decoded by a self-splitting Gaussian head to predict Gaussian primitive parameters.

During training, we replace standard 3DGS rasterization \cite{kerbl3DGS} with stochastic solid volume rendering. This formulation models Gaussian primitives as continuous stochastic solids and renders them along camera rays \cite{2023Objectsasvolumes}. It improves the quality of synthesized Gaussian depth maps and strengthens geometric reconstruction accuracy. Unlike existing feed-forward 3DGS methods that typically train the full network, our framework freezes the original visual foundation model, including the backbone, depth head, and camera head. Only the multi-scale voxel module and Gaussian primitive decoding head are optimized. This design substantially reduces training cost and allows our method to be plugged into patch-feature-based visual foundation models.

\begin{enumerate}
    \item We propose $\text{VG}^2$GT, a feed-forward, non-pixel-aligned 3DGS reconstruction framework for geometrically accurate Gaussian scene reconstruction from an arbitrary number of unposed and uncalibrated images.
    
    \item We introduce a multi-scale voxel-based feature enhancement strategy that splits and regresses Gaussian primitives from fine voxel features. Combined with stochastic solid volume rendering supervision, this design reduces overlapping artifacts and geometric errors commonly observed in pixel-aligned feed-forward 3DGS methods.
    
    \item Extensive experiments on multiple datasets show that $\text{VG}^2$GT achieves state-of-the-art geometric reconstruction accuracy and novel view synthesis quality compared with existing feed-forward 3DGS methods.

    \item $\text{VG}^2$GT substantially reduces the training cost, enabling high-quality 3DGS reconstruction on frozen visual foundation models using only academic-scale computational resources, while remaining plug-and-play to any visual foundation models.
\end{enumerate}

\section{Related Works}

\subsection{Traditional Pipelines}
Traditional pipelines for geometric reconstruction and novel view synthesis (NVS) from RGB images typically consist of three stages: camera pose estimation, dense reconstruction, and per-scene optimization. Camera poses and sparse 3D points can be estimated using hand-crafted features \cite{sift}, learned descriptors \cite{detone2018superpoint}, and feature matching methods \cite{sarlin2020superglue}. Given the estimated poses and dense image features, dense depth estimation can then be performed \cite{colmapmvs, yao2018mvsnet}. Plane sweeping has been widely adopted in this stage because it balances computational cost and reconstruction performance \cite{plane-sweeping}. After dense point-cloud reconstruction, traditional pipelines usually refine the scene representation through per-scene optimization. Neural radiance fields \cite{mildenhall2020nerf} and 3D Gaussian Splatting \cite{kerbl3DGS} are representative examples for improving reconstruction quality and enabling NVS.

Although each component has advanced substantially, the decoupled design of traditional pipelines can lead to error accumulation. Errors introduced during camera pose estimation may propagate to subsequent per-scene optimization and degrade NVS quality \cite{zybapin2025robust}. The separated stages also increase computational cost and reduce overall efficiency.

\subsection{Optimization-based Novel View Synthesis}
Common 3D scene representations include meshes, point clouds, voxel grids, multi-plane images, and neural implicit functions. In recent years, radiance-field methods have shown strong potential for NVS and 3D reconstruction. Neural Radiance Fields (NeRF) and its variants \cite{mildenhall2020nerf,barron2022mipnerf360} map 3D positions and viewing directions to colors and densities using multi-layer perceptrons (MLPs), enabling high-quality NVS. To improve computational efficiency and transferability, 3D Gaussian Splatting (3DGS) \cite{kerbl3DGS} represents a scene as explicit Gaussian primitives and enables real-time NVS through differentiable rasterization. Subsequent works have improved rendering quality \cite{scaffoldgs}, geometric accuracy \cite{Huang2DGS2024,huang2025fatesgs,zhang2026gggs}, and training efficiency \cite{ren2025fastgs}.

Most NeRF and 3DGS methods require accurate camera parameters, which are usually estimated by classical SfM methods \cite{colmapsfm} or improved variants \cite{glomap,wang2024vggsfm}. This dependency can introduce additional errors. Several recent works therefore jointly optimize camera parameters and scene representations. However, these methods either require sequential RGB-D frames with known intrinsics \cite{keetha2024splatam,gsicpslam}, or are limited to restricted scene settings \cite{meng2021gnerf,wang2021nerf--}. They also require per-scene training. In contrast, our method directly regresses 3DGS scenes through a feed-forward network, avoiding camera-parameter estimation and per-scene optimization.

\subsection{Visual Foundation Models}
With advances in parallel computing and large-scale neural networks, visual foundation models (VFMs) have shown strong potential for 3D vision. DUSt3R \cite{dust3r_cvpr24} performs end-to-end reconstruction from two unconstrained RGB images to point maps using self-attention and cross-attention. VGGT \cite{wang2025vggt} further introduces alternating attention to support joint reasoning over multi-view RGB images. Subsequent works have explored improved encoder designs \cite{keetha2026mapanything}, training strategies \cite{depthanything3}, and permutation equivariance across input views \cite{wang2025pi3}. These models show that camera estimation, depth estimation, and point-map prediction can be learned jointly with a shared large-scale Transformer backbone.

Existing VFMs commonly use ViT-based encoders, such as DINO \cite{oquab2023dinov2}, together with alternating attention backbones. Sparse outputs, including camera parameters, are typically predicted by MLP heads. Dense outputs, including depth maps and point maps, are commonly decoded by DPT \cite{DPT}. Through DPT, patch tokens that are usually downsampled by a factor of 14 can be progressively upsampled and aligned with image pixels, enabling full-resolution depth and point-cloud estimation.

Several works have extended VFMs to Gaussian scene regression \cite{xu2024freesplatter,depthanything3}. However, most methods decode Gaussian scenes with DPT heads, producing pixel-aligned Gaussian primitives. This formulation often leads to non-uniform Gaussian distributions and multi-view inconsistencies \cite{li2026tokensplat}. AnySplat \cite{jiang2025anysplat} performs non-pixel-aligned Gaussian reasoning through differentiable voxelization. However, it still first regresses pixel-aligned Gaussian primitives and then aggregates them into a Gaussian scene. Our method directly decodes Gaussian primitives from structured voxel features, avoiding this intermediate pixel-aligned representation.

\section{Method}

\begin{figure*}[pos=t]
\centering
\includegraphics[width = 0.99\linewidth]{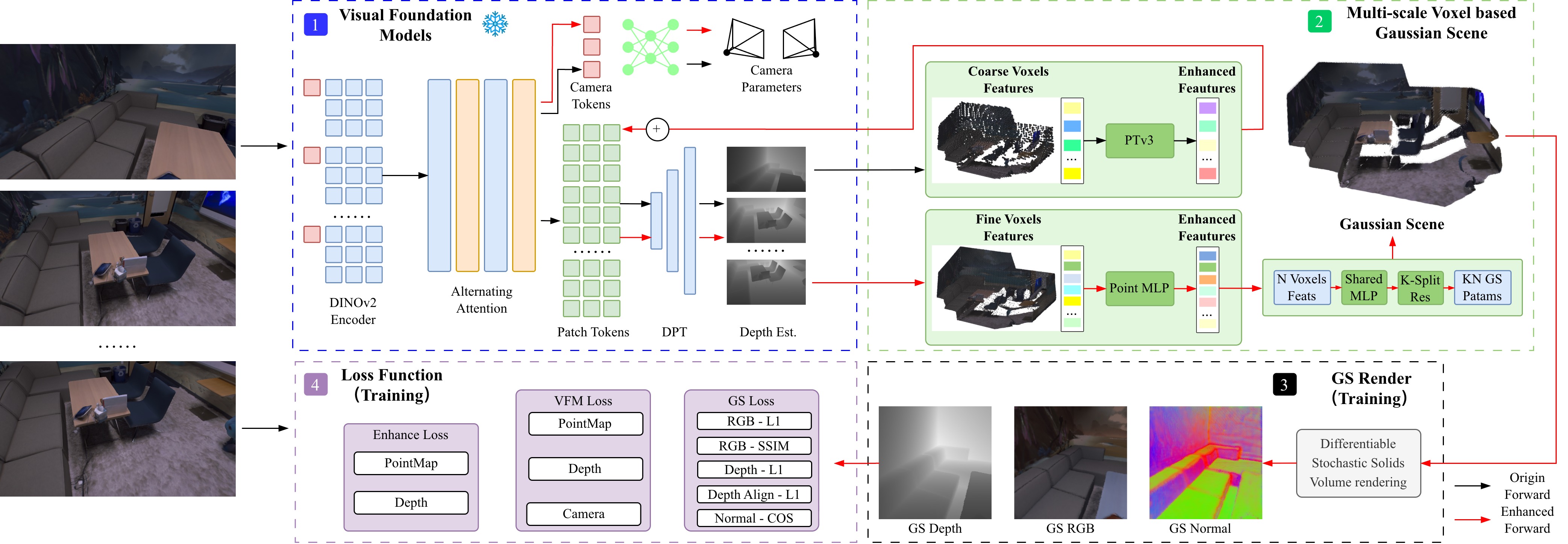}
\caption{Overview of $\text{VG}^2$GT. The frozen VFM takes multi-view RGB images as input, computes patch tokens, and decodes depth maps and camera parameters. Multi-scale voxels decode the Gaussian scene: coarse voxels enhance patch tokens, while fine voxels split and decode Gaussian primitives. During training, differentiable stochastic solid volume rendering supervises scene geometry.}\label{fig_full_method}
\end{figure*}

We propose $\text{VG}^2$GT, a feed-forward framework for constructing 3DGS representations from unconstrained RGB images. Built on VGGT \cite{wang2025vggt}, our method improves global scene consistency through multi-scale voxel module. This design improves both novel view synthesis (NVS) quality and geometric reconstruction accuracy, as illustrated in \autoref{fig_full_method}.

\subsection{Preliminaries and Bottlenecks}
\subsubsection{Preliminaries}
Feed-forward 3D reconstruction methods, such as VGGT \cite{wang2025vggt} and DUSt3R \cite{dust3r_cvpr24}, recover 3D scene representations from $N$ unposed images $\left\{\mathbf{I}_{i}\right\}_{i=1}^{N}$. A large visual Transformer $\mathcal{F}$ jointly encodes and interprets the input images. With a shared alternating-attention backbone and independent prediction heads, it produces pixel-aligned point maps $\mathbf{P}_{i} \in \mathbb{R}^{H\times W\times 3}$, depth maps $\mathbf{D}_{i} \in \mathbb{R}^{H\times W\times 1}$, and camera pose representations $\mathbf{C}_{i}\in \mathbb{R}^9$, as formulated in \autoref{eqn_preliminary_vggt}.
\begin{equation}
\label{eqn_preliminary_vggt}
\mathcal{F}_{\phi}\left(\left\{\mathbf{I}_{i}\right\}_{i=1}^{N}\right)=\left\{\mathbf{P}_{i},\mathbf{D}_{i},\mathbf{C}_{i}\right\}_{i=1}^{N}
\end{equation}

To enhance the scene representation and enable NVS, 3DGS \cite{kerbl3DGS} represents a scene as a set of Gaussian primitives. Each primitive defines a 3D Gaussian distribution parameterized by a center $\mu_{g} \in \mathbb{R}^3$, opacity $\sigma_{g}\in \mathbb{R}^+$, rotation quaternion $r_{g}\in \mathbb{R}^4$, scale $s_{g}\in \mathbb{R}^3$, and color represented by $k$-th order spherical harmonics $c_{g}\in \mathbb{R}^{3\times(k+1)^2}$. Given the target camera pose $\mathbf{C}_{i}$, 3DGS projects each 3D Gaussian onto the image plane as a 2D Gaussian. It then renders the image using depth-ordered $\alpha$-blending, as shown in \autoref{eqn_3dgs_render_preliminary}. Here, $\Pi(\cdot)$ denotes Gaussian projection under the camera pose, $\left(\mu_{g}^{(i)},\Sigma_{g}^{(i)}\right)$ denote the projected pixel coordinate and covariance of the $g$-th Gaussian in the $i$-th view, and $o_{g}^{(i)}(\mathbf{p})$ denotes its effective opacity at pixel $\mathbf{p}$.
\begin{equation}
\label{eqn_3dgs_render_preliminary}
\begin{aligned}
\left(\mu_{g}^{(i)},\Sigma_{g}^{(i)}\right)&=\Pi\left(\mathbf{C}_{i};\mu_{g},r_{g},s_{g}\right)\\
\mathbf{I}_{i}(\mathbf{p})&=\sum_{g\in\mathcal{G}(\mathbf{p})} c_{g}\,o_{g}^{(i)}(\mathbf{p})\prod_{j<g}\left(1-o_{j}^{(i)}(\mathbf{p})\right)
\end{aligned}
\end{equation}

\subsubsection{Bottlenecks}
VGGT obtains one token for each image patch, typically corresponding to a $14\times14$ pixel region. The resulting image features are therefore downsampled by a factor of 14. To recover dense depth maps, VGGT uses DPT \cite{DPT} to progressively upsample image tokens into pixel-aligned features. To obtain feed-forward 3DGS scenes, representative methods such as Depth Anything 3 (DA3) \cite{depthanything3} apply a DPT head on the backbone features. A lightweight MLP then regresses Gaussian attributes for each pixel.

However, pixel-aligned features lack global consistency in feed-forward networks and can introduce excessive overlap, since the architecture enforces a one-to-one correspondence between pixels and Gaussian primitives. We instead construct a non-pixel-aligned feed-forward network $\mathcal{N}_\theta$ that jointly predicts a 3DGS scene $\mathbf{G}$ and camera parameters $\left\{\mathbf{C}_{i}\right\}_{i=1}^{N}$ from $N$ unposed images $\left\{\mathbf{I}_{i}\right\}_{i=1}^{N}$, as formulated in \autoref{eqn_propose}.
\begin{equation}
	\label{eqn_propose}
	\mathcal{N}_\theta(\left\{\mathbf{I}_{i}\right\}_{i=1}^{N}) = \left\{\left(\mu_{g}, \sigma_{g}, r_{g}, s_{g}, c_{g}\right)\right\}_{g=1}^{G}\cup \left\{\mathbf{C}_{i}\right\}_{i=1}^{N}
\end{equation}

Here, the 3DGS scene $\mathbf{G}$ consists of $G$ non-pixel-aligned Gaussian primitives. The number of primitives is not tied one-to-one to pixels or views, and therefore grows much more slowly than in pixel-aligned methods.

\subsection{Differentiable Multi-scale Voxel Module}
Our network backbone follows VGGT \cite{wang2025vggt}. It uses DINOv2 \cite{oquab2023dinov2} to extract multi-view image features and an alternating-attention backbone for scene understanding. Camera parameters and depth maps are predicted jointly.

\subsubsection{Coarse-level Feature Enhancement}
On top of this backbone, we introduce multi-scale voxel features to build a compact spatial representation. Given the predicted depth maps $\left\{\mathbf{D}_{i}\right\}_{i=1}^{N}$ and camera parameters $\left\{\mathbf{C}_{i}\right\}_{i=1}^{N}$, each pixel is projected into 3D space to obtain dense pixel-aligned global point clouds $\left\{\mathbf{P}_{i}\right\}_{i=1}^{N}$. We also extract the undecoded image patch features $\left\{\mathbf{F}_{i}^P\right\}_{i=1}^{N}$ from the backbone. Following related voxel-based designs \cite{zhu2025voxelsplat,wang2025amb3r}, we align global point clouds with image patch features by interpolation and point-cloud downsampling, and then construct global voxel features $\mathbf{F}^v$. Each voxel feature $\mathbf{f}_m^v$ is computed as a confidence-weighted average over all 3D points $\Omega_m$ inside the voxel, where the weights are given by depth confidence $\left\{\mathbf{N}^D_{i}\right\}_{i=1}^{N}$, as shown in \autoref{eqn_voxelfeature}. The confidence of each voxel is defined as the minimum point confidence within the voxel, reducing the influence of unreliable points on reconstruction.
\begin{equation}
	\label{eqn_voxelfeature}
	\begin{aligned}
\mathbf{f}_m^v &= \frac{\sum_{{p} \in \Omega_m} \mathbf{n}(\mathbf{p})\,\mathbf{f}({p})}{\sum_{{p} \in \Omega_m} \mathbf{n}({p})}\\
\mathbf{n}_m^v &= \min_{{p} \in \Omega_m} \mathbf{n}({p})
	\end{aligned}
\end{equation}

We then use PointTransformerV3 (PTv3) \cite{wu2024ptv3} to enhance each voxel feature. PTv3 jointly models spatial and feature distributions with a permutation-invariant serialized point-cloud network. Specifically, we serialize the sparse voxel grid into a one-dimensional feature sequence with $\mathcal{S}$, enhance the sequence with PTv3 $\mathcal{T}_{\psi}$, and deserialize the enhanced features back into voxel space, as shown in \autoref{eqn_ptvzengqiang}.
\begin{equation}
	\label{eqn_ptvzengqiang}
\left\{\hat{\mathbf{f}}_m^v\right\}_{m=1}^{M} = \left(\mathcal{S}^{-1} \circ \mathcal{T}_{\psi} \circ \mathcal{S}\right)\left(\left\{\mathbf{f}_m^v\right\}_{m=1}^{M}\right),
\end{equation}

After obtaining the enhanced voxel features, we map them back to the 3D points in $\Omega_m$ for subsequent processing. For each point $\mathbf{p}$, we find its 16 nearest voxel centers with k-nearest neighbors (KNN). The enhanced voxel features are then interpolated back to the point by distance-weighted aggregation.
This process produces enhanced features aligned with image patches. We fuse these features into backbone patch features using zero convolution, and then decode depth maps and camera parameters again. Although this design requires two decoding rounds, most VFM runtime lies in the backbone. The decoding heads add only marginal overhead, so the overall framework remains efficient.

\subsubsection{Fine-level Gaussian Scene Decoding}
Next, we repeat the computation in \autoref{eqn_voxelfeature} to obtain fine voxel features from the enhanced image patches.
To capture scene geometry at a finer granularity, we no longer downsample the global point cloud. Instead, following the design of the depth head, we use a DPT head to upsample image features and align them with the global point cloud.

After obtaining fine-grained voxel features $\mathbf{f}_m^{dv}$, we use PointMLP \cite{pointmlp} instead of PTv3 for feature enhancement. PTv3 has substantially higher memory consumption when processing million-point-scale point clouds, which can lead to GPU out-of-memory errors. In contrast, PointMLP is more memory efficient and stable for large-scale point clouds because of its MLP-based permutation-invariant architecture, as shown in \autoref{eqn_pointmlp}, where $\mathcal{M}_{\phi}$ denotes the PointMLP network.
\begin{equation}
	\label{eqn_pointmlp}
\left\{\hat{\mathbf{f}}_m^{dv}\right\}_{m=1}^{M} = \left(\mathcal{S}^{-1} \circ \mathcal{M}_{\phi} \circ \mathcal{S}\right)\left(\left\{\mathbf{f}_m^{dv}\right\}_{m=1}^{M}\right),
\end{equation}

Given the dense voxel features $\left\{\hat{\mathbf{f}}_m^{dv}\right\}$ and voxel centers $\left\{{\mathbf{x}}_m^{dv}\right\}$, we decode the Gaussian scene $\mathbf{G}$. To stabilize training, we use residual predictions for the Gaussian center $\mu_g$ and color $c_g$. The initial color of each voxel is computed as a weighted average of image colors. Spherical harmonic coefficients above order 0 are initialized to 0, and the initial Gaussian center is set to the voxel center. The network then predicts residual offsets for the Gaussian center and spherical harmonic coefficients. The Gaussian scale $s_g$ is obtained by applying a sigmoid activation with an upper bound of twice the voxel size. The rotation quaternion $r_g$ and opacity $\sigma_g$ are directly regressed by MLPs.

Voxel sparsity inevitably limits the recoverable scene detail. To improve fine-grained accuracy, we introduce a self-splitting design in which each voxel feature regresses multiple sets of Gaussian primitive parameters. Together with residual center prediction, this design allows Gaussian primitives within a single voxel to preserve local geometric complexity.

\subsection{Continuous Stochastic Solids Volume Rendering}
To further improve the geometric accuracy of 3DGS, we replace standard 3DGS rasterization \cite{kerbl3DGS} with volume rendering. We treat Gaussian primitives as continuous stochastic solids and render them along camera rays \cite{2023Objectsasvolumes}, as shown in \autoref{eqn_volume_render}.
\begin{equation}
\label{eqn_volume_render}
\begin{aligned}
\mathbf{I}_{\mathrm{pix}} &= \int_{t_n}^{t_f} p(t)\,\mathbf{c}(\mathbf{x}(t),\omega)\,dt=\mathbf{c}\left(1-\mathbf{v}\left(t\right)^{2}\right),\\
p(t) &= T(t)\,\sigma(\mathbf{x}(t),\omega),\\
T(t) &= \exp\left(-\int_{t_n}^{t} \sigma(\mathbf{x}(s),\omega)\,ds\right)
\end{aligned}
\end{equation}
Here, $\mathbf{x}(t)$ denotes the 3D position along the camera ray at depth $t$, $\omega$ denotes the viewing direction, $T(t)$ denotes accumulated transmittance, and $\mathbf{v}$ denotes the vacancy of the solid.

\begin{figure}[pos=ht]
\centering
\includegraphics[width=0.9\linewidth]{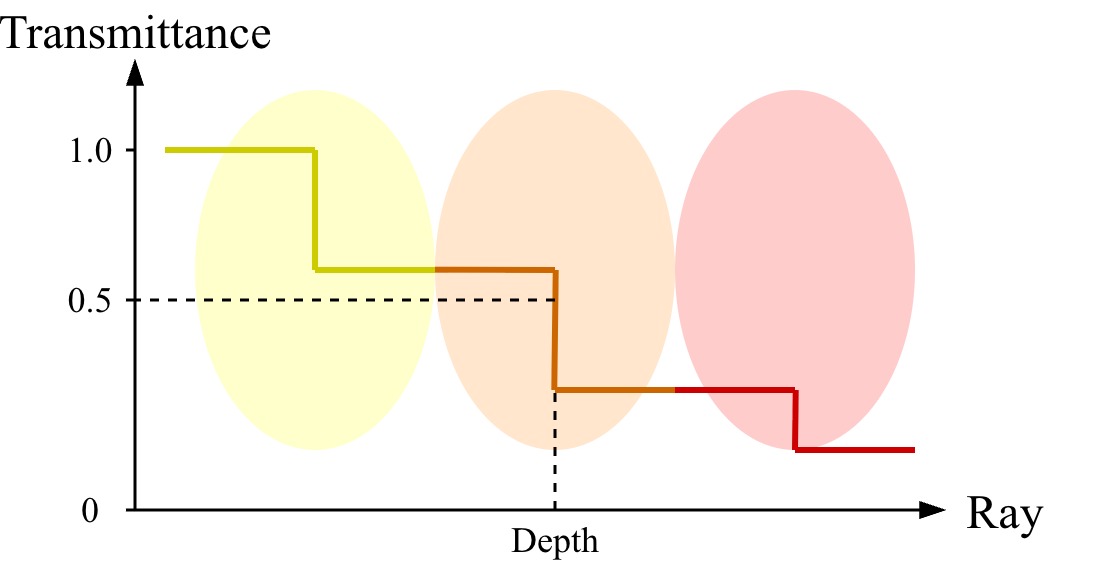}\\
(a) Rasterized depth rendering of 3DGS
\includegraphics[width=0.9\linewidth]{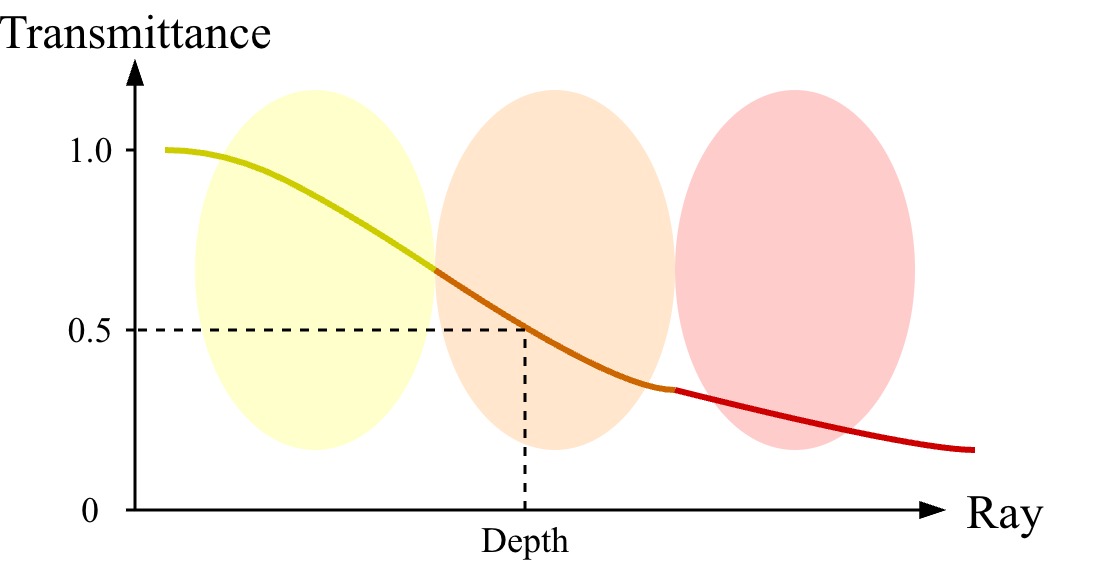}\\
(b) Stochastic solid volume rendering of our method
\caption{Comparison of different depth rendering strategies. Compared with rasterization-based median depth, our continuous volume rendering produces smoother and more accurate depth maps.}
\label{fig_depth_render_compare}
\end{figure}

Because our feed-forward process also predicts depth maps, the rendered depth of Gaussian primitives can be aligned with the VFM-regressed depth maps. Existing works improve the geometric quality of 3DGS scenes by adding regularization, constraining Gaussian shapes, or rendering depth maps with rasterization-based strategies \cite{Huang2DGS2024,huang2025fatesgs}. However, $\alpha$-blending-based depth rendering mixes foreground and background depths, which limits geometric accuracy. Median-depth rasterization avoids this averaging, but its gradients are propagated only to the selected median primitive, and it can also produce abrupt depth changes near primitive boundaries. We instead model Gaussian primitives as continuous entities.
To make volume rendering equivalent to rasterization, we compute the corresponding vacancy $\mathbf{v}$. Following \cite{zhang2026gggs}, color rendering can be expressed as \autoref{eqn_gggs_color}.
\begin{equation}
\label{eqn_gggs_color}
\mathbf{I}_{\mathrm{pix}} = \mathbf{c}G(t^*)
\end{equation}
Here, $t^*$ denotes the location of the maximum Gaussian response along the ray. Combining \autoref{eqn_volume_render} and \autoref{eqn_gggs_color} gives the vacancy formulation in \autoref{eqn_vacancy}. This formulation makes volume rendering equivalent to rasterization while allowing transmittance to vary continuously along each ray. As shown in \autoref{fig_depth_render_compare}, this continuous formulation improves depth synthesis quality.
\begin{equation}
\label{eqn_vacancy}
\mathbf{v} = \sqrt{1-{G(t^*)}}
\end{equation}

During forward propagation, we define the depth map as the median depth, namely the location where accumulated transmittance reaches 0.5. The accumulated transmittance of multiple Gaussian primitives is computed by multiplicative composition. The median location is then solved with a fixed number of bisection iterations. During backward propagation, gradients are propagated with the closed-form derivative of depth with respect to Gaussian parameters, as shown in \autoref{eqn_tmed_grad}.
\begin{equation}
\label{eqn_tmed_grad}
\frac{\partial d_{}}{\partial \theta} = -\frac{\partial T\left(d;\theta\right)}{\partial \theta} \bigg/ \frac{\partial T\left(d;\theta\right)}{\partial d}
\end{equation}
Here, $d$ denotes the median depth location of a pixel, and $\theta$ denotes the parameters of Gaussian primitives.

Through this pipeline, we obtain a fully differentiable process from unposed images $\left\{\mathbf{I}_{i}\right\}_{i=1}^{N}$ to the Gaussian scene $\mathbf{G}$, and then to rendered images $\left\{\mathbf{I}_{i}^\mathbf{G}\right\}_{i=1}^{N}$ and depth maps $\left\{\mathbf{D}_{i}^\mathbf{G}\right\}_{i=1}^{N}$.
\begin{figure}[pos=htbp]
\centering
\includegraphics[width = 0.99\linewidth]{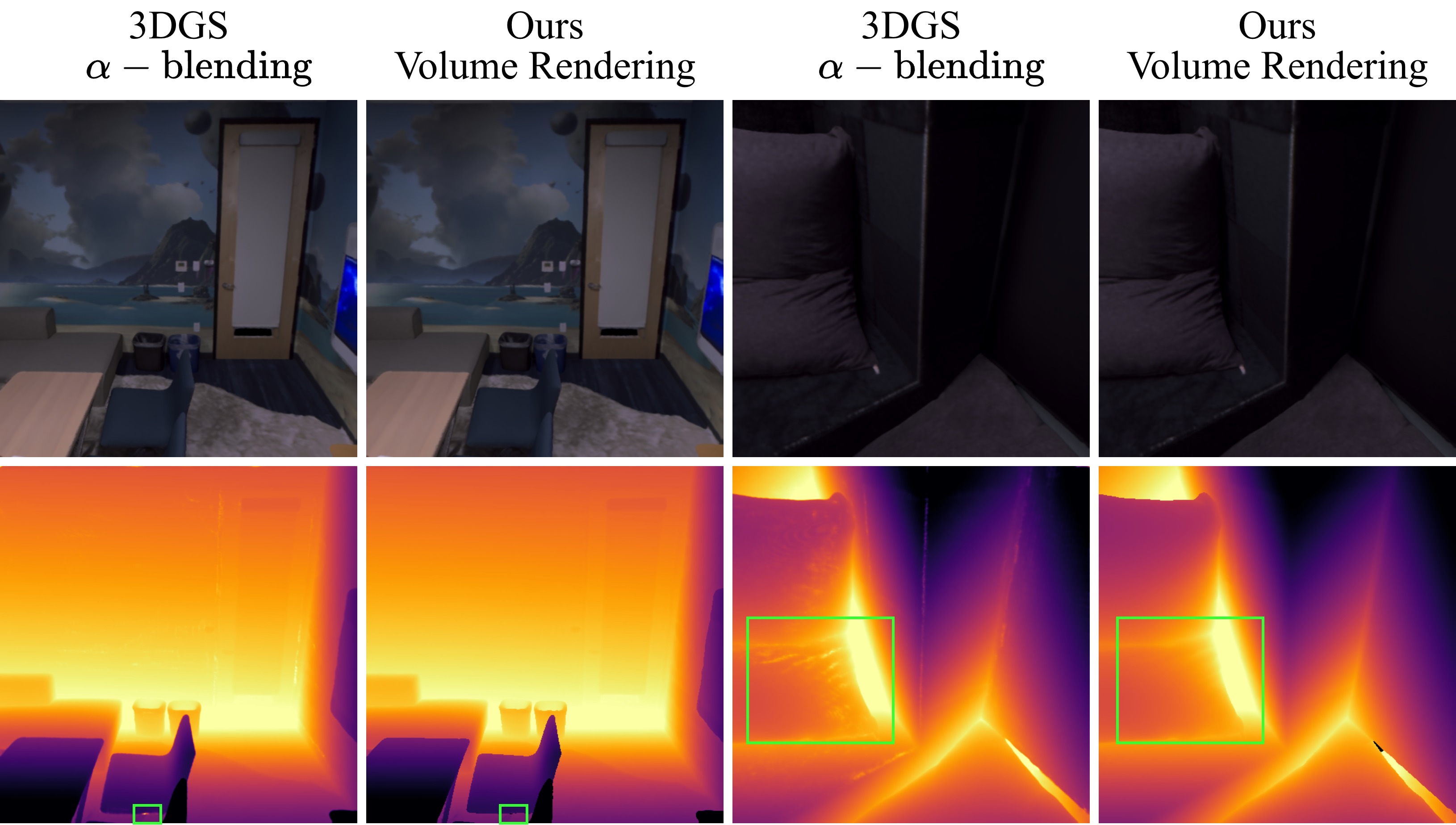}
\caption{RGB images and depth maps rendered by rasterization and volume rendering for the same Gaussian scene.}\label{fig_sup_volumerender_good}
\end{figure}

As shown in \autoref{fig_sup_volumerender_good}, we render the same Gaussian scene using vanilla rasterization and continuous stochastic solids volume rendering. The resulting RGB images are nearly identical, whereas the depth maps differ noticeably. Vanilla rasterization produces floating-depth artifacts, while volume rendering recovers more accurate depth details. The ablation studies and quantitative results further support the effectiveness of volume rendering for improving geometric reconstruction accuracy.

\subsection{Training}
We use DA3-GIANT \cite{depthanything3} as the VFM for training. Throughout training, the DA3 backbone and all original decoders are kept frozen. We discard the original DA3 Gaussian head and train only the proposed differentiable multi-scale voxel module.
This design allows our method to be plugged into arbitrary feed-forward patch-feature-based VFMs while preserving the representation capability of the pretrained backbone.

\textbf{Datasets}. Following prior works \cite{li2025iggt,wang2025vggt}, we train our model on a mixture of three datasets: the synthetic Infinigen dataset, the video-captured KITTI dataset \cite{kittidepth,kittiraw}, and the RGB-D ScanNet dataset \cite{dai2017scannet}. These datasets cover indoor and outdoor scenes under both synthetic and real-captured conditions. This diversity improves model generalization and transferability to unseen scenes.

\textbf{Loss Functions}. We first adopt the same training losses as VGGT, supervising point maps with $\mathcal{L}_\text{pointmap}$ and depth maps with $\mathcal{L}_\text{depth}$.
To improve training stability, we also record the losses before image patch feature enhancement, including the point-map loss $\hat{\mathcal{L}}_\text{pointmap}$ and depth loss $\hat{\mathcal{L}}_\text{depth}$.
We introduce an additional supervision term that encourages loss reduction after enhancement:
\begin{equation}
\label{eqn_enhance_loss}
\begin{aligned}
\mathcal{L}_\text{*}^\text{enhance} &= \lambda_\text{*}^\text{enhance} \cdot \max\left(0, \hat{\mathcal{L}}_\text{*} - \mathcal{L}_\text{*}\right)\\
\mathcal{L}_\text{enhance} &= \mathcal{L}_\text{pointmap}^\text{enhance} + \mathcal{L}_\text{depth}^\text{enhance} 
\end{aligned}
\end{equation}
With $\mathcal{L}_\text{enhance}$, the enhanced decoded outputs are explicitly encouraged to improve over the pre-enhancement outputs. This stabilizes the enhancement module while preserving knowledge from the original VFM.

We then supervise the Gaussian scene generated through volume rendering, as shown in \autoref{eqn_gsloss}. Following per-scene 3DGS training, we compute the L1 loss $\mathcal{L}_1^c$ and SSIM loss \cite{wang2004ssim} $\mathcal{L}_\text{SSIM}^c$ on rendered RGB images. To strengthen geometric reconstruction, we additionally align rendered depth maps with the depth maps regressed by the feed-forward network. We also apply an absolute loss between rendered depth maps and ground-truth depth. These terms jointly improve rendered-depth accuracy and consistency with decoded depth maps.
\begin{equation}
\label{eqn_gsloss}
\begin{aligned}
\mathcal{L}_\mathbf{G} &= \mathcal{L}_1^c + \mathcal{L}_\text{SSIM}^c + \mathcal{L}_1(\mathbf{D},\mathbf{D}_\mathbf{G}) + \mathcal{L}_1(\mathbf{D}_\text{gt},\mathbf{D}_\mathbf{G}) + \mathcal{L}_n
\end{aligned}
\end{equation}

To encourage feed-forward Gaussian primitives to align with scene surfaces, we introduce a normal regularization loss following 2DGS \cite{Huang2DGS2024}. Unlike 2DGS, which derives rendered normal maps from rendered depth maps, we render normal maps directly from Gaussian primitives. This decouples normal supervision from depth rendering. For each Gaussian primitive, we compute the tangent-plane normal of the density surface at the ray intersection under the current viewing direction. We then render the normal map using the same volume rendering formulation as RGB rendering. The normal loss $\mathcal{L}_n$ is computed by cosine similarity between the rendered normal map and the normal map derived from the depth map.

\textbf{Implementation Details}. 
We optimize the proposed objective with the AdamW optimizer for 420K iterations. The initial learning rate is set to $3\times10^{-6}$ and decayed to $1\times10^{-8}$ using a cosine annealing schedule. Following prior work \cite{jiang2025anysplat}, we set the default voxel size to $0.002$ meters, split each voxel feature into two Gaussian primitives, and use an image resolution of $224\times448$. At each iteration, we randomly sample 3-18 images. Training is conducted on two NVIDIA RTX PRO 6000 GPUs and takes 49.5 hours, corresponding to 99 GPU hours.

\section{Experiments}

\providecommand{\thickhline}{\noalign{\global\arrayrulewidth=0.8pt}\hline\noalign{\global\arrayrulewidth=0.4pt}}
\begin{table*}[pos=ht]
\centering
\footnotesize
\renewcommand{\arraystretch}{0.99}
\begin{tabularx}{\textwidth}{>{\raggedright\arraybackslash}p{1.90cm}|*{4}{>{\centering\arraybackslash}X}|*{4}{>{\centering\arraybackslash}X}|*{4}{>{\centering\arraybackslash}X}}
\thickhline
\multirow{2}{*}{\textbf{Method}} &
\multicolumn{4}{c|}{\textbf{3 Views}} &
\multicolumn{4}{c|}{\textbf{10 Views}} &
\multicolumn{4}{c}{\textbf{20 Views}} \\
& RMSE$\downarrow$ & ABS$\downarrow$ & $\delta_{1.25}\uparrow$ & $\delta_{1.05}\uparrow$
& RMSE$\downarrow$ & ABS$\downarrow$ & $\delta_{1.25}\uparrow$ & $\delta_{1.05}\uparrow$
& RMSE$\downarrow$ & ABS$\downarrow$ & $\delta_{1.25}\uparrow$ & $\delta_{1.05}\uparrow$ \\ \thickhline
\multicolumn{13}{l}{\textit{Replica Dataset\cite{replica19arxiv}}} \\
MVSplat\cite{chen2024mvsplat}        & 0.547 & 0.143 & 0.790 & 0.272 & 0.501 & 0.114 & 0.872 & 0.474 & 0.780 & 0.251 & 0.715 & 0.336 \\
VolSplat\cite{wang2025volsplat}       & 0.698 & 0.191 & 0.683 & 0.165 & 0.407 & 0.099 & 0.903 & 0.346 & 0.461 & 0.130 & 0.853 & 0.229 \\
DepthSplat\cite{xu2024depthsplat}     & 0.279 & 0.059 & 0.942 & 0.707 & 0.279 & 0.053 & 0.954 & 0.744 & 0.180 & 0.040 & 0.969 & 0.818 \\ \hline
FreeSplatter\cite{xu2024freesplatter}   & 0.797 & 0.284 & 0.559 & 0.144 & 0.918 & 0.270 & 0.569 & 0.138 & 0.619 & 0.205 & 0.677 & 0.191 \\
VGGT\cite{wang2025vggt}           & 0.973 & 0.263 & 0.458 & 0.111 & 0.972 & 0.244 & 0.500 & 0.122 & 0.792 & 0.219 & 0.563 & 0.142 \\
AnySplat\cite{jiang2025anysplat}       & 0.156 & 0.039 & 0.992 & 0.710 & 0.171 & 0.041 & 0.990 & 0.751 & 0.121 & 0.036 & 0.995 & 0.757 \\
YoNoSplat\cite{ye2025yonosplatneedmodelfeedforward}      & 0.285 & 0.080 & 0.947 & 0.434 & 0.237 & 0.059 & 0.972 & 0.632 & 0.155 & 0.047 & 0.985 & 0.674 \\
DA3\cite{depthanything3}            & 0.619 & 0.192 & 0.670 & 0.180 & 0.394 & 0.116 & 0.878 & 0.289 & 0.296 & 0.092 & 0.922 & 0.364 \\
\textbf{$\text{VG}^2$GT(3DGS)} & 0.092 & 0.011 & 0.995 & 0.940 & 0.109 & 0.014 & 0.996 & 0.980 & 0.100 & 0.022 & 0.980 & 0.966 \\
\textbf{$\text{VG}^2$GT}  & \textbf{0.069} & \textbf{0.007} & \textbf{0.997} & \textbf{0.990} & \textbf{0.078} & \textbf{0.011} & \textbf{0.997} & \textbf{0.990} & \textbf{0.043} & \textbf{0.006} & \textbf{0.999} & \textbf{0.994} \\ \thickhline
\multicolumn{13}{l}{\textit{DTU Dataset\cite{dtu}}} \\
MVSplat\cite{chen2024mvsplat}        & 0.069 & 0.663 & 0.163 & 0.037 & 0.092 & 0.093 & 0.905 & 0.430 & 0.140 & 0.129 & 0.836 & 0.398 \\
VolSplat\cite{wang2025volsplat}       & 0.033 & 0.382 & 0.515 & 0.124 & 0.080 & 0.087 & 0.895 & 0.481 & 0.097 & 0.094 & 0.904 & 0.476 \\
DepthSplat\cite{xu2024depthsplat}     & 0.022 & 0.284 & 0.748 & 0.261 & 0.070 & 0.089 & 0.928 & 0.388 & 0.105 & 0.103 & 0.924 & 0.386 \\ \hline
FreeSplatter\cite{xu2024freesplatter}   & 0.030 & 0.333 & 0.702 & 0.320 & 0.092 & 0.116 & 0.839 & 0.346 & 0.090 & 0.113 & 0.848 & 0.343 \\
VGGT\cite{wang2025vggt}           & 0.031 & 0.338 & 0.555 & 0.141 & 0.123 & 0.149 & 0.778 & 0.179 & 0.118 & 0.143 & 0.781 & 0.209 \\
AnySplat\cite{jiang2025anysplat}       & 0.016 & 0.259 & 0.961 & 0.587 & 0.031 & 0.026 & 0.988 & 0.892 & 0.023 & 0.019 & \textbf{0.993} & 0.935 \\
YoNoSplat\cite{ye2025yonosplatneedmodelfeedforward}      & 0.033 & 0.441 & 0.433 & 0.105 & 0.124 & 0.164 & 0.696 & 0.205 & 0.143 & 0.196 & 0.615 & 0.170 \\
DA3\cite{depthanything3}            & 0.029 & 0.398 & 0.625 & 0.162 & 0.037 & 0.034 & 0.982 & 0.831 & 0.027 & 0.022 & 0.991 & 0.906 \\
\textbf{$\text{VG}^2$GT(3DGS)} & \textbf{0.014} & \textbf{0.237} & \textbf{0.962} & \textbf{0.759} & 0.026 & 0.020 & \textbf{0.988} & 0.935 & 0.021 & 0.016 & 0.992 & 0.950 \\
\textbf{$\text{VG}^2$GT}  & 0.015 & 0.238 & 0.961 & 0.757 & \textbf{0.026} & \textbf{0.019} & 0.988 & \textbf{0.937} & \textbf{0.021} & \textbf{0.015} & 0.992 & \textbf{0.954} \\ \thickhline
\multicolumn{13}{l}{\textit{TAT Dataset\cite{TAT}}} \\
MVSplat\cite{chen2024mvsplat}        & 0.839 & 0.483 & 0.291 & 0.065 & 1.003 & 0.776 & 0.109 & 0.024 & 0.835 & 0.436 & 0.353 & 0.106 \\
VolSplat\cite{wang2025volsplat}       & 0.673 & 0.355 & 0.556 & 0.164 & 0.348 & 0.321 & 0.498 & 0.138 & 0.418 & 0.200 & 0.693 & 0.234 \\
DepthSplat\cite{xu2024depthsplat}     & 0.488 & 0.295 & 0.723 & 0.351 & 0.431 & 0.291 & 0.655 & 0.229 & 0.413 & 0.179 & 0.733 & 0.354 \\ \hline
FreeSplatter\cite{xu2024freesplatter}   & 0.536 & 0.344 & 0.686 & 0.234 & 0.507 & 0.491 & 0.551 & 0.164 & 0.486 & 0.252 & 0.567 & 0.136 \\
VGGT\cite{wang2025vggt}           & 0.629 & 0.292 & 0.477 & 0.113 & 0.400 & 0.284 & 0.514 & 0.118 & 0.611 & 0.290 & 0.494 & 0.101 \\
AnySplat\cite{jiang2025anysplat}       & 0.394 & 0.089 & 0.936 & 0.592 & 0.192 & 0.093 & 0.919 & 0.606 & 0.371 & 0.129 & 0.866 & 0.629 \\
YoNoSplat\cite{ye2025yonosplatneedmodelfeedforward}      & 0.570 & 0.221 & 0.721 & 0.236 & 0.327 & 0.266 & 0.612 & 0.201 & 0.452 & 0.315 & 0.574 & 0.145 \\
DA3\cite{depthanything3}            & 0.571 & 0.263 & 0.567 & 0.161 & 0.293 & 0.315 & 0.555 & 0.144 & 0.415 & 0.224 & 0.654 & 0.212 \\
\textbf{$\text{VG}^2$GT(3DGS)} & \textbf{0.381} & \textbf{0.077} & \textbf{0.947} & \textbf{0.740} & \textbf{0.145} & \textbf{0.057} & 0.950 & \textbf{0.777} & \textbf{0.264} & \textbf{0.075} & 0.921 & \textbf{0.741} \\
\textbf{$\text{VG}^2$GT}  & 0.389 & 0.081 & 0.947 & 0.733 & 0.148 & 0.058 & \textbf{0.952} & 0.773 & 0.265 & 0.077 & \textbf{0.926} & 0.708 \\ \thickhline
\end{tabularx}
\vspace{-1mm}
\caption{Quantitative Gaussian-scene depth rendering results on Replica, DTU, and TAT under multiple input-view settings.}\label{tbl_depth_whole}
\label{tab:depth_results}
\vspace{-5mm}
\end{table*}

\begin{figure*}[pos=htbp]
\centering
\includegraphics[width = 0.99\linewidth]{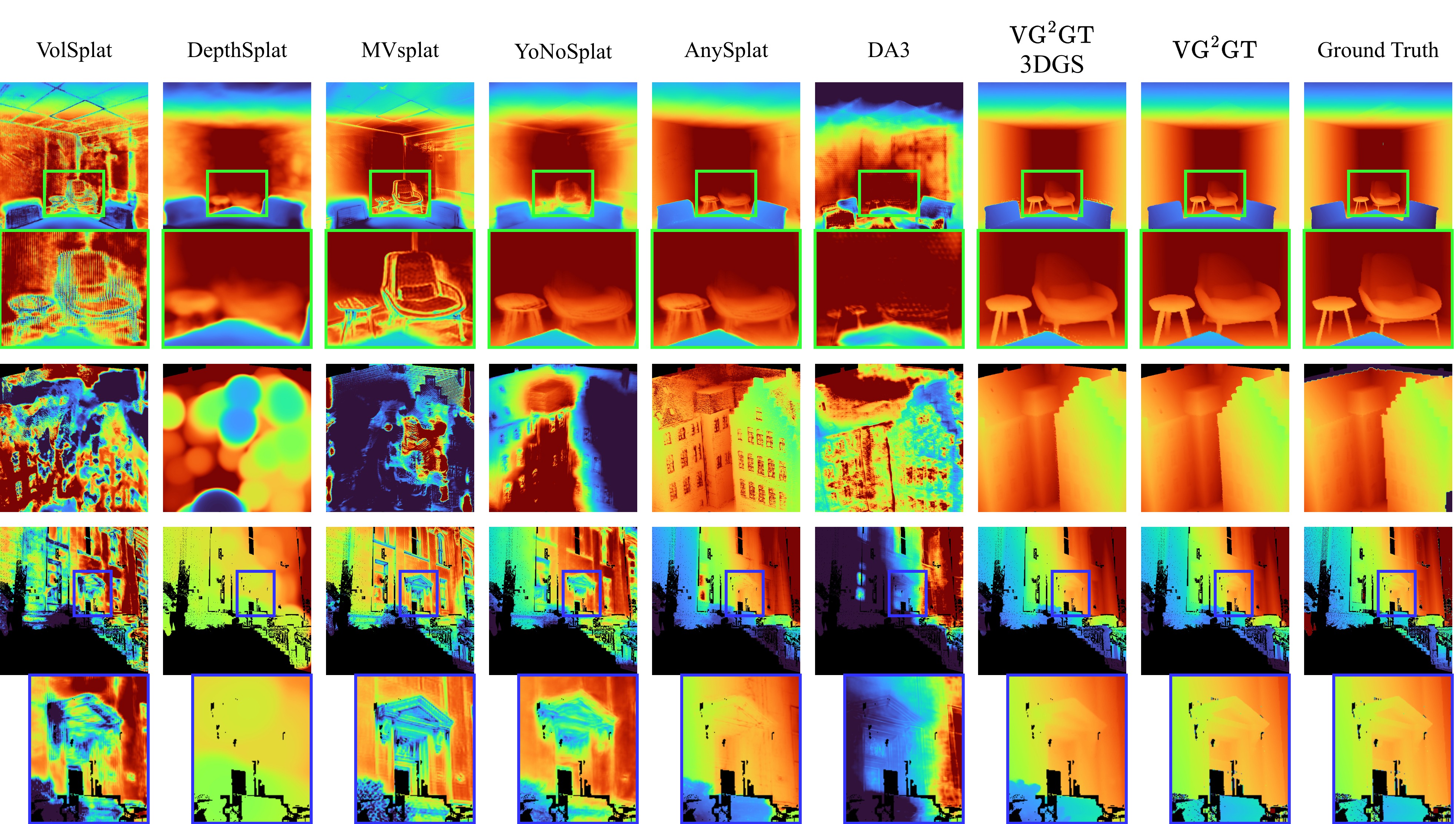}
\vspace{-2mm}
\caption{Visual comparisons between $\text{VG}^2$GT and baselines for geometric reconstruction. Gaussian depth maps rendered by $\text{VG}^2$GT show higher accuracy, indicating improved geometric fidelity of the reconstructed Gaussian scenes.}\label{fig_exp_depth}
\end{figure*}

\begin{table*}[pos=ht]
\centering
\footnotesize
\renewcommand{\arraystretch}{1.02}
\begin{tabularx}{0.99\textwidth}{>{\raggedright\arraybackslash}p{1.95cm}|*{3}{>{\centering\arraybackslash}X}|*{3}{>{\centering\arraybackslash}X}|*{3}{>{\centering\arraybackslash}X}}
\thickhline
\multirow{2}{*}{\textbf{Method}} &
\multicolumn{3}{c|}{\textbf{3 Views}} &
\multicolumn{3}{c|}{\textbf{10 Views}} &
\multicolumn{3}{c}{\textbf{20 Views}} \\
& PSNR$\uparrow$ & SSIM$\uparrow$ & LPIPS$\downarrow$
& PSNR$\uparrow$ & SSIM$\uparrow$ & LPIPS$\downarrow$
& PSNR$\uparrow$ & SSIM$\uparrow$ & LPIPS$\downarrow$ \\ \thickhline
\multicolumn{10}{l}{\textit{Replica Dataset\cite{replica19arxiv}}} \\
MVSplat\cite{chen2024mvsplat}        & 25.70 & 0.830 & 0.240 & 24.18 & 0.835 & 0.257 & 24.17 & 0.807 & 0.300 \\
VolSplat\cite{wang2025volsplat}      & 19.26 & 0.681 & 0.453 & 28.66 & 0.876 & 0.180 & 22.17 & 0.747 & 0.400 \\
DepthSplat\cite{xu2024depthsplat}    & 22.98 & 0.798 & 0.403 & 23.64 & 0.797 & 0.420 & 23.55 & 0.788 & 0.419 \\ \hline
FreeSplatter\cite{xu2024freesplatter} & 18.19 & 0.707 & 0.510 & 16.78 & 0.653 & 0.540 & 16.85 & 0.699 & 0.530 \\
VGGT\cite{wang2025vggt}              & 13.61 & 0.608 & 0.626 & 14.80 & 0.582 & 0.589 & 14.89 & 0.595 & 0.579 \\
AnySplat\cite{jiang2025anysplat}     & 27.56 & \textbf{0.877} & \textbf{0.151} & 25.76 & 0.843 & 0.208 & 25.40 & 0.837 & 0.205 \\
YoNoSplat\cite{ye2025yonosplatneedmodelfeedforward} & 28.48 & 0.871 & 0.239 & 29.06 & 0.884 & 0.241 & 28.76 & 0.878 & 0.239 \\
DA3\cite{depthanything3}             & 21.38 & 0.747 & 0.257 & 23.83 & 0.804 & 0.243 & 25.00 & 0.812 & 0.213 \\
\textbf{$\text{VG}^2$GT(3DGS)}       & 29.87 & 0.876 & 0.184 & 30.08 & 0.883 & 0.208 & 29.54 & 0.876 & 0.202 \\
\textbf{$\text{VG}^2$GT}             & \textbf{31.03} & \textbf{0.898} & 0.169 & \textbf{30.87} & \textbf{0.898} & 0.191 & \textbf{30.41} & \textbf{0.894} & \textbf{0.184} \\ \thickhline
\multicolumn{10}{l}{\textit{TAT Dataset\cite{TAT}}} \\
MVSplat\cite{chen2024mvsplat}        & 19.83 & 0.628 & 0.359 & 18.06 & 0.545 & 0.441 & 14.62 & 0.412 & 0.584 \\
VolSplat\cite{wang2025volsplat}      & 16.67 & 0.527 & 0.474 & 17.78 & 0.519 & 0.485 & 16.77 & 0.454 & 0.537 \\
DepthSplat\cite{xu2024depthsplat}    & 16.90 & 0.477 & 0.679 & 15.21 & 0.446 & 0.675 & 14.59 & 0.420 & 0.699 \\ \hline
FreeSplatter\cite{xu2024freesplatter} & 15.30 & 0.455 & 0.608 & 12.92 & 0.419 & 0.693 & 12.49 & 0.408 & 0.715 \\
VGGT\cite{wang2025vggt}              & 14.89 & 0.456 & 0.615 & 14.15 & 0.444 & 0.650 & 13.99 & 0.434 & 0.675 \\
AnySplat\cite{jiang2025anysplat}     & 20.75 & 0.676 & 0.285 & 17.60 & 0.590 & 0.347 & 16.90 & 0.547 & 0.392 \\
YoNoSplat\cite{ye2025yonosplatneedmodelfeedforward} & 20.44 & 0.572 & 0.418 & 19.82 & 0.552 & 0.413 & 19.49 & 0.528 & 0.443 \\
DA3\cite{depthanything3}             & 18.22 & 0.483 & 0.379 & 18.98 & 0.519 & 0.348 & 18.89 & 0.504 & \textbf{0.355} \\
\textbf{$\text{VG}^2$GT(3DGS)}       & 21.68 & 0.691 & 0.294 & 19.89 & 0.618 & 0.328 & 19.34 & 0.563 & 0.393 \\
\textbf{$\text{VG}^2$GT}             & \textbf{22.46} & \textbf{0.716} & \textbf{0.272} & \textbf{20.22} & \textbf{0.650} & \textbf{0.321} & \textbf{20.01} & \textbf{0.588} & 0.383 \\ \thickhline
\end{tabularx}
\caption{Quantitative NVS results on the Replica and TAT datasets under multiple input-view settings.}
\label{tbl_nvs_whole}
\end{table*}

\begin{figure*}[pos=htbp]
\centering
\includegraphics[width = 0.99\linewidth]{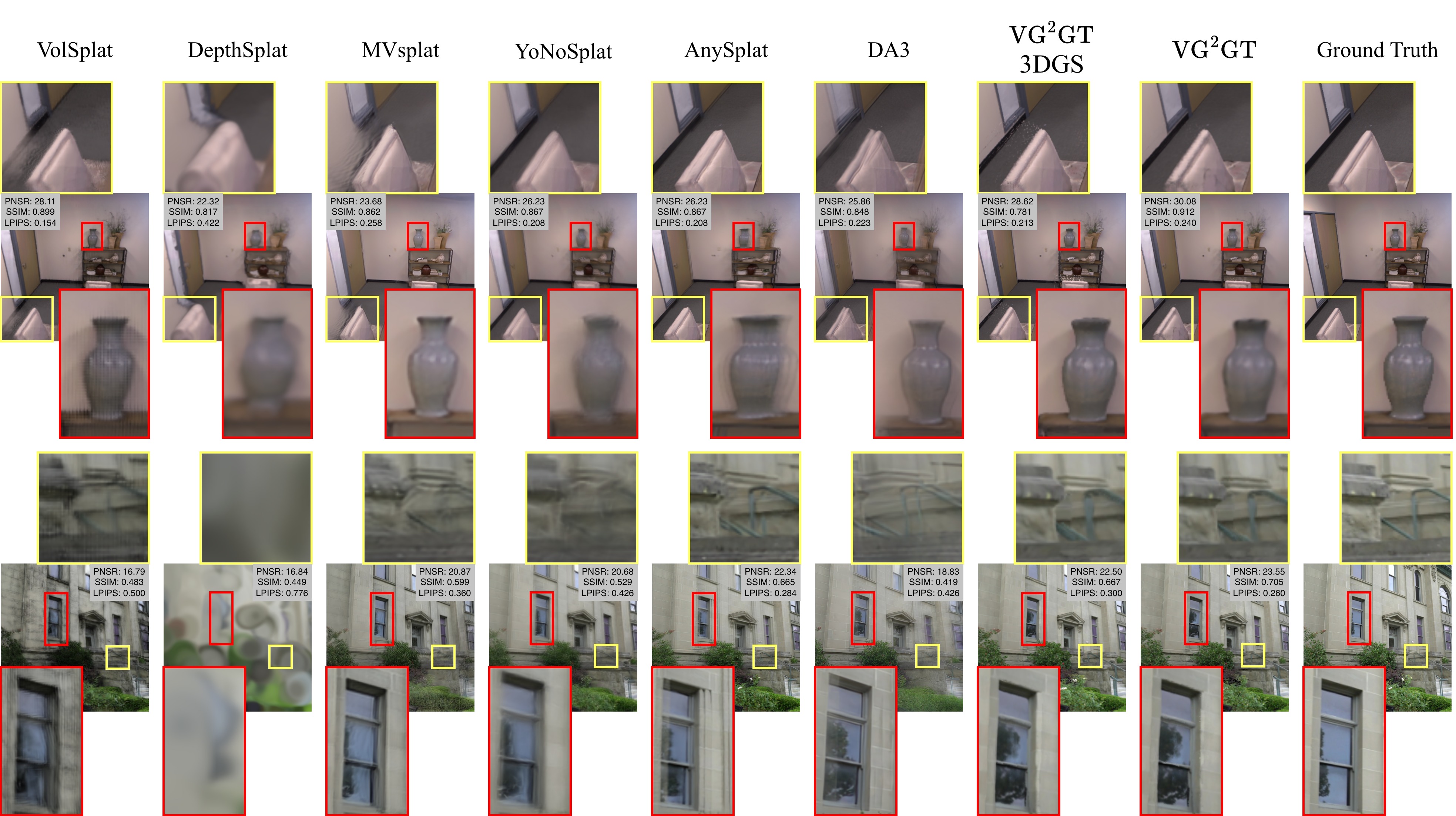}
\vspace{-2mm}
\caption{Visual comparisons between $\text{VG}^2$GT and baselines for NVS. Multi-scale voxel-based Gaussian scenes suppress the overlapping artifacts commonly observed in pixel-aligned methods, improving scene geometry and NVS quality.}\label{fig_exp_NVS}
\end{figure*}

\subsection{Experimental Setup}
\textbf{Metrics.} Following prior works \cite{wang2025vggt,depthanything3,jiang2025anysplat}, we evaluate both the reconstruction quality and NVS performance of the Gaussian scenes produced by $\text{VG}^2$GT. For geometric reconstruction, we report RMSE (L2 error), ABS (L1 error), and $\delta_n$ accuracy for rendered depth maps, as defined in \autoref{eq_delta_n}.
\begin{equation}
\label{eq_delta_n}
    \max\left(\frac{\hat{D}_i}{D_i}, \frac{D_i}{\hat{D}_i}\right) < n
\end{equation}
where $\hat{D}_i$ and $D_i$ denote the predicted and ground-truth depths at pixel $i$. For rendered normal maps, we report mean angular error (Mean$^\circ$), median angular error (Median$^\circ$), thresholded angular accuracy (Acc@10$^\circ$ and Acc@30$^\circ$), and cosine similarity (Cos.). For novel view synthesis, we evaluate rendering fidelity using PSNR, SSIM \cite{SSIM}, and LPIPS \cite{LPIPS}.

We further evaluate the spatial uniformity of the reconstructed scene point cloud using the coefficient of variation (CV). For each point, we identify its $N$ nearest neighbours with KNN and compute the mean neighbour distance. CV is defined as the ratio between the standard deviation and the mean of these per-point distances, as shown in \autoref{eqn_junyun_cv}. A lower CV indicates a more uniform point-cloud distribution.
\begin{equation}
  \label{eqn_junyun_cv}
  \mathrm{CV}
=
\frac{
\sqrt{
\frac{1}{N}
\sum_{i=1}^{N}
\left(
d_i - \frac{1}{N}\sum_{i=1}^{N} d_i
\right)^2
}
}{
\frac{1}{N}\sum_{i=1}^{N} d_i
}
\end{equation}

\noindent\textbf{Datasets.}
We evaluate all methods on three datasets: Replica \cite{replica19arxiv}, which contains synthetic indoor scenes; DTU \cite{dtu}, a standard benchmark for geometry reconstruction from captured RGB images; and Tanks and Temples (TAT) \cite{TAT}, which contains real outdoor scenes. This protocol covers diverse scene types and allows robustness to be assessed across different environments. During evaluation, all input images are center-cropped and resized to $518\times518$.

\noindent\textbf{Baselines.}
We compare against state-of-the-art (SOTA) pose-free feed-forward 3DGS methods, including FreeSplatter \cite{xu2024freesplatter}, AnySplat \cite{jiang2025anysplat}, YoNoSplat \cite{ye2025yonosplatneedmodelfeedforward}, and DA3 \cite{depthanything3}. These methods predict pixel-aligned Gaussian scenes, while AnySplat further introduces a voxel module to re-voxelize and aggregate the predicted Gaussians. For a fair comparison, we use the same default voxel size as our method ($0.002\,\mathrm{m}$).
We also include posed feed-forward 3DGS baselines, including MVSplat \cite{chen2024mvsplat}, VolSplat \cite{wang2025volsplat}, and DepthSplat \cite{xu2024depthsplat}. In addition, we compare with the standard 3DGS initialization strategy \cite{kerbl3DGS}, where VGGT \cite{wang2025vggt} generates colored point clouds for initializing the Gaussian scene.

\subsection{Evaluation of Reconstruction Geometry}
We evaluate geometric reconstruction on the DTU, Replica, and TAT datasets. To address the scale ambiguity of pose-free reconstruction, we follow prior works \cite{wang2025amb3r,jiang2025anysplat} and first globally align each predicted depth map to the ground-truth depth map. RMSE, ABS, $\delta_{1.25}$, and $\delta_{1.05}$ are then computed on the aligned depth maps.
For DTU and TAT, the ground-truth depth maps are obtained by back-projecting scanned point clouds. We therefore apply valid depth masks before both alignment and evaluation.

We evaluate geometric reconstruction under different numbers of input views. For each scene, we randomly select 3, 10, and 20 input images. During training, RGB images and depth maps are rendered from the Gaussian scene using stochastic solid volume rendering. During evaluation, the generated Gaussian scenes are rendered using both the vanilla 3DGS rasterizer \cite{kerbl3DGS} and stochastic solid volume rendering.
As shown in \autoref{tbl_depth_whole} and \autoref{fig_exp_depth}, the Gaussian scenes regressed by our method achieve higher geometric accuracy than state-of-the-art baselines across most datasets and input-view settings.
Pixel-aligned Gaussian scene regression methods, represented by DA3, often produce overlapping artifacts when multi-view Gaussian primitives are misaligned by coupled depth and camera-pose errors. These artifacts degrade scene geometry. In contrast, the proposed differentiable multi-scale voxel module, combined with stochastic solid volume rendering supervision, suppresses these artifacts and improves geometric accuracy.

\subsection{Evaluation of Novel View Synthesis}
We evaluate novel view synthesis on the Replica and TAT datasets. For each scene, we also select 3, 10, and 20 input images and report PSNR, SSIM, and LPIPS. As shown in \autoref{tbl_nvs_whole} and \autoref{fig_exp_NVS}, our method consistently outperforms state-of-the-art baselines across most input-view settings. By suppressing overlapping artifacts, $\text{VG}^2$GT improves scene geometry and thereby enhances novel view synthesis performance.

\subsection{Evaluation of Normal Map}
\begin{figure}[pos=htbp]
\centering
\includegraphics[width = 0.99\linewidth]{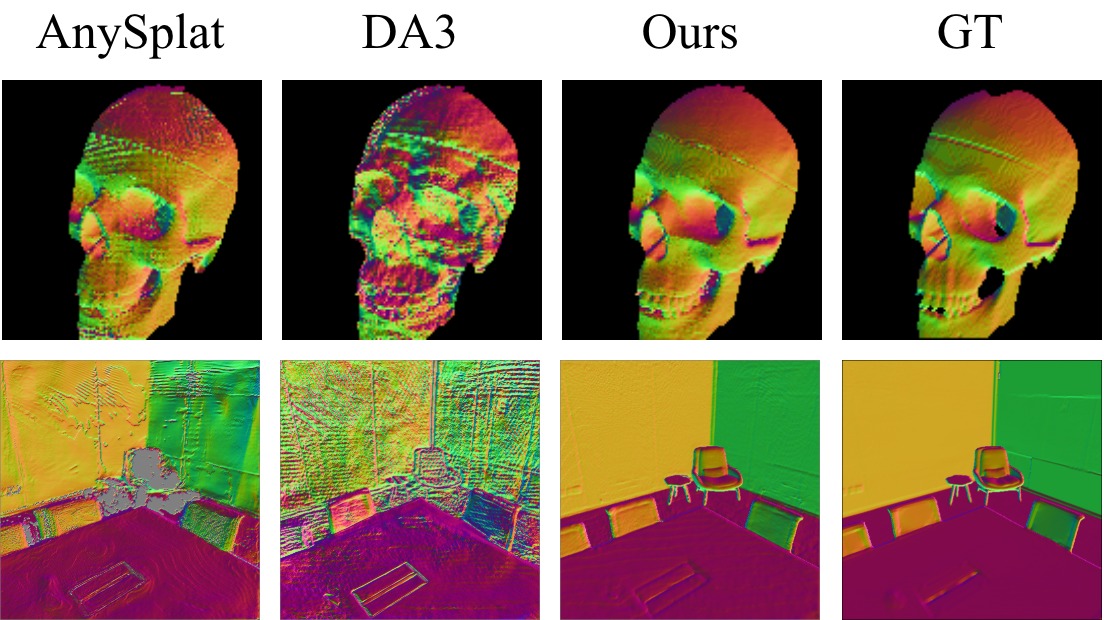}
\vspace{-2mm}
\caption{Visual comparisons of rendered normal maps. $\text{VG}^2$GT produces more accurate and spatially consistent normal maps than pixel-aligned baselines.}\label{fig_exp_normal}
\end{figure}

$\text{VG}^2$GT's accurate Gaussian scene geometry also improves the quality of rendered normal maps. As shown in \autoref{tbl_camera_orientation} and \autoref{fig_exp_normal}, $\text{VG}^2$GT achieves clear advantages in normal-map rendering on the Replica and DTU datasets.

\begin{table}[pos=ht]
\centering
\scriptsize
\renewcommand{\arraystretch}{1.05}
\begin{tabularx}{0.49\textwidth}{>{\raggedright\arraybackslash}p{1.35cm}|*{5}{>{\centering\arraybackslash}X}}
\thickhline
\textbf{Method} & {Mean$^\circ$$\downarrow$} & {Median$^\circ$$\downarrow$} & {Acc@10$^\circ$$\uparrow$} & {Acc@30$^\circ$$\uparrow$} & {Cos.$\uparrow$} \\ \thickhline
\multicolumn{6}{l}{\textit{Replica Dataset\cite{replica19arxiv}}} \\
YoNoSplat  & 59.279 & 61.548 & 0.020 & 0.154 & 0.461 \\
AnySplat   & 25.035 & 14.958 & 0.360 & 0.724 & 0.831 \\
DA3        & 52.797 & 52.102 & 0.056 & 0.268 & 0.534 \\
\textbf{$\text{VG}^2$GT} & \textbf{7.339} & \textbf{3.636} & \textbf{0.839} & \textbf{0.955} & \textbf{0.972} \\ \thickhline
\multicolumn{6}{l}{\textit{DTU Dataset\cite{dtu}}} \\
YoNoSplat  & 44.400 & 42.100 & 0.050 & 0.345 & 0.654 \\
AnySplat   & 26.087 & 20.837 & 0.193 & 0.691 & 0.850 \\
DA3        & 60.484 & 62.141 & 0.016 & 0.136 & 0.445 \\
\textbf{$\text{VG}^2$GT} & \textbf{18.607} & \textbf{14.077} & \textbf{0.337} & \textbf{0.842} & \textbf{0.914} \\ \thickhline
\end{tabularx}
\caption{Quantitative normal-map accuracy on the Replica and DTU datasets.}
\label{tbl_camera_orientation}
\end{table}

\subsection{Evaluation of Time Consumption}

The training and inference costs of $\text{VG}^2$GT remain well controlled. We evaluate the time consumption of $\text{VG}^2$GT using 10 input images.
\begin{figure}[pos=htbp]
\centering
\begin{subfigure}{0.98\linewidth}
    \centering
    \includegraphics[width=\linewidth]{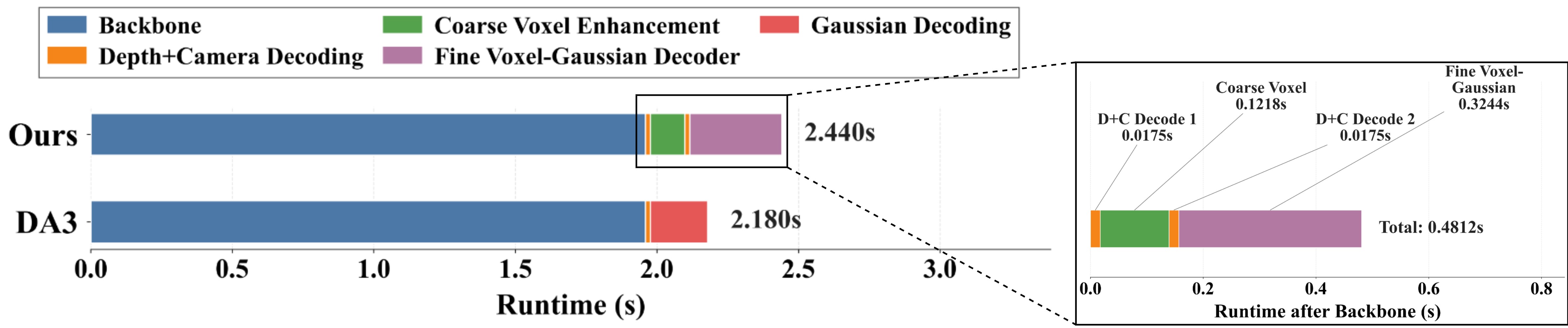}
    \caption{Inference time.}
    \label{fig_time_infer}
\end{subfigure}
\vspace{1mm}
\begin{subfigure}{0.98\linewidth}
    \centering
    \includegraphics[width=\linewidth]{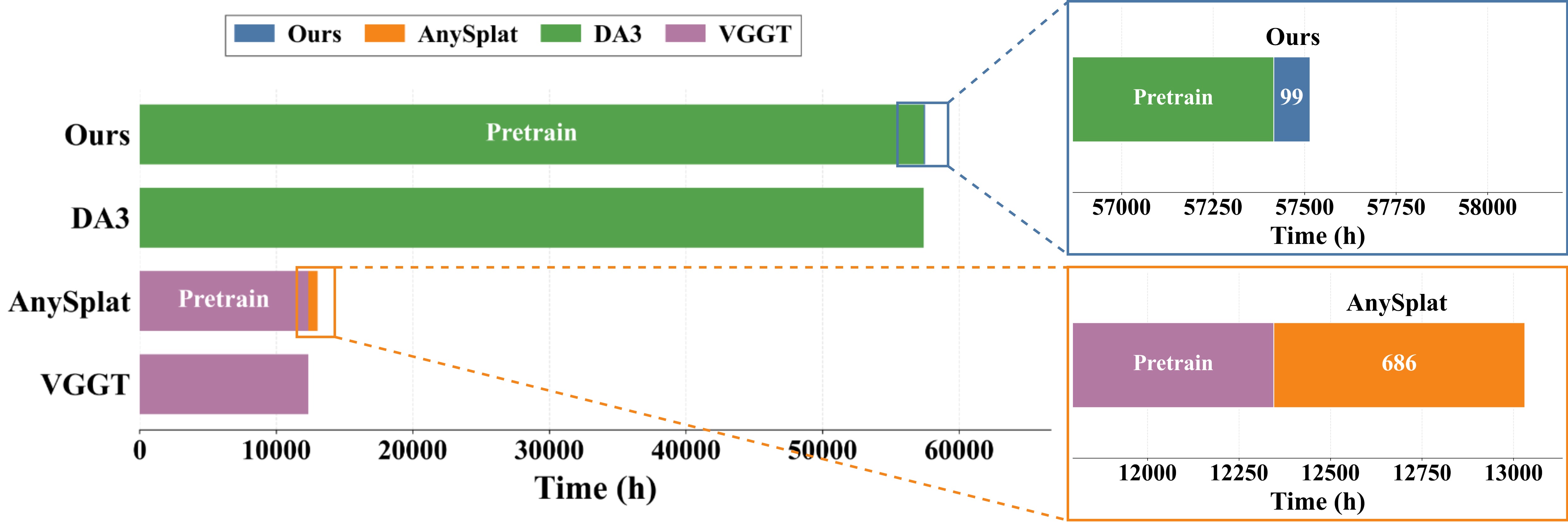}
    \caption{Training time.}
    \label{fig_time_train}
\end{subfigure}
\caption{Time consumption comparisons for inference and training.}
\label{fig_time_consumption}
\end{figure}

\textbf{Inference}. Although $\text{VG}^2$GT decodes depth maps and camera parameters twice, the additional decoding overhead is small relative to the backbone. It requires only 0.0852 s at a resolution of $224\times448$.
Compared with the original DA3 (0.764 s), our method increases inference time by 0.358 s, reaching 1.122 s in total. Most of the additional cost comes from fine-level Gaussian scene decoding (0.293 s). Notably, owing to our voxel-based Gaussian scene decoding, when the image resolution increases to $518\times518$, the fine-voxel decoding time remains almost unchanged. The total inference time is 2.440 s, only 0.260 s higher than DA3 (2.180 s), as shown in \autoref{fig_time_infer}.
\begin{figure}[pos=htbp]
\centering
\includegraphics[width = 0.99\linewidth]{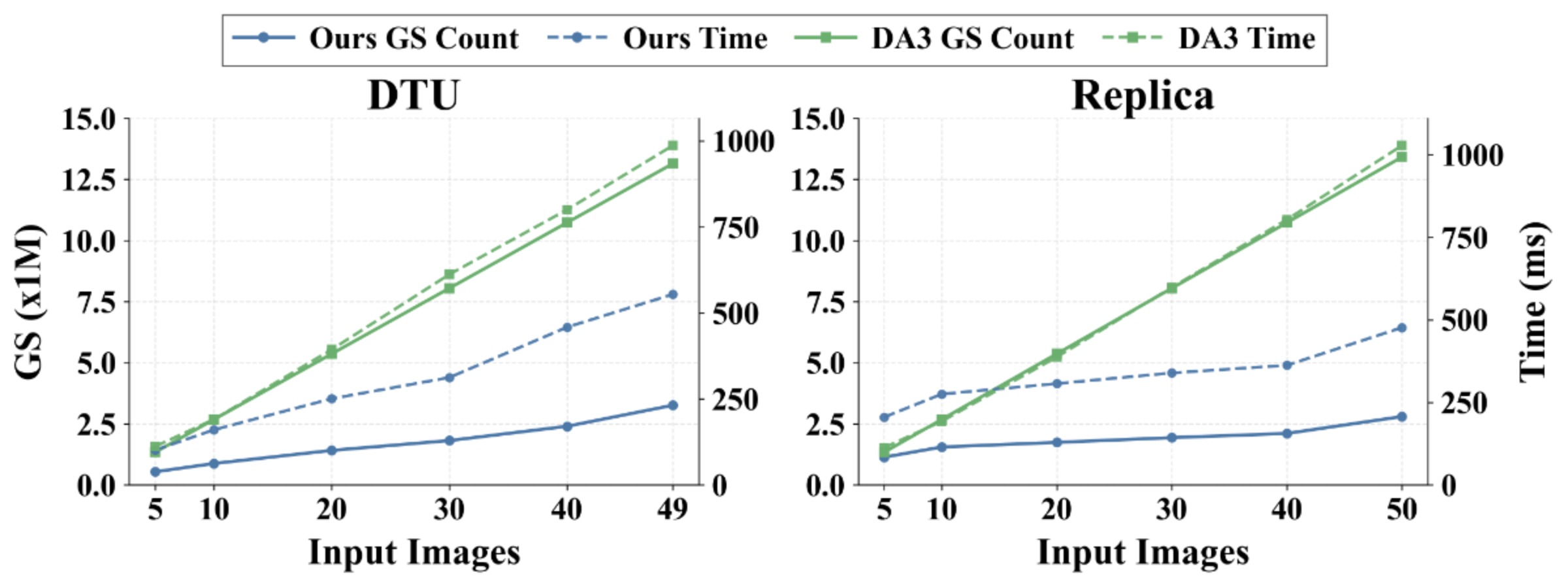}
\vspace{-2mm}
\caption{Number of Gaussian primitives and decoding-head time as the number of input images increases.}\label{fig_Voxel_GS_count}
\end{figure}

Voxel-based alignment also controls both the number of Gaussian primitives and the decoding cost as the number of input views increases. As shown in \autoref{fig_Voxel_GS_count}, the pixel-aligned baseline DA3 shows an approximately linear increase in both primitive count and decoding time. In contrast, $\text{VG}^2$GT uses multi-scale voxelization to fuse features in overlapping regions as additional views are introduced. With approximately 50 input images, the decoding time and primitive count are reduced to 24.83\% and 56.2\% of DA3 on DTU, and to 20.88\% and 46.38\% on Replica.
In addition, \autoref{tbl_point_uniformity} shows that the point cloud generated by $\text{VG}^2$GT achieves a much lower CV than DA3 and reaches a level comparable to AnySplat based on voxel-aggregation. These results indicate that voxel-based method significantly improves point-cloud uniformity.

\begin{table}[pos=ht]
\centering
\footnotesize
\renewcommand{\arraystretch}{1.05}
\begin{tabularx}{0.49\textwidth}{>{\raggedright\arraybackslash}p{1.1cm}|*{4}{>{\centering\arraybackslash}X}}
\thickhline
 & \textbf{YoNoSplat} & \textbf{AnySplat} & \textbf{DA3} & \textbf{$\text{VG}^2$GT} \\ \thickhline
CV$\downarrow$ & 0.644 & 0.387 & 7.642 & \textbf{0.353} \\ \thickhline
\end{tabularx}
\caption{Point-cloud uniformity evaluation on the Replica dataset.}
\label{tbl_point_uniformity}
\end{table}

\textbf{Training}. Compared with other pixel-aligned 3DGS methods (AnySplat: 768 A800 GPU hours; DA3: 30,720 H100 GPU hours) and VFMs (VGGT: 13,824 A100 GPU hours; MapAnything: 9,216 H200 GPU hours), our method requires substantially lower training cost, using only 99 RTX PRO 6000 GPU hours. This efficiency mainly comes from keeping the VFM fully frozen during training. As shown in \autoref{fig_time_train} and \autoref{tbl_training_cost}, using the time conversion of \cite{dao2023flashattention2}, our method reduces the required training time compared with backbone-finetuning strategies represented by AnySplat, while supporting plug-and-play adaptation to VFMs.

\begin{table}[pos=ht]
\centering
\footnotesize
\renewcommand{\arraystretch}{1.05}
\begin{tabularx}{0.49\textwidth}{>{\raggedright\arraybackslash}p{1.45cm}
                                  >{\centering\arraybackslash}X
                                  >{\centering\arraybackslash}X
                                  >{\centering\arraybackslash}X}
\thickhline
\textbf{Method} & \textbf{GPU} & \textbf{Training Time(h)} & \textbf{Converted Time(h)} \\ \thickhline
Ours        & PRO 6000 & 99    & 99 \\
AnySplat    & A800         & 768   & 685.824 \\
DA3         & H100         & 30720 & 57415.68 \\
VGGT        & A100         & 13824 & 12344.832 \\
MapAnything & H200         & 9216  & 24689.664 \\ \thickhline
\end{tabularx}
\caption{Training-cost comparison across different methods.}
\label{tbl_training_cost}
\end{table}

\subsection{Evaluation of Coarse-level Feature Enhancement}
Coarse-level feature enhancement improves both subsequent fine-level Gaussian scene decoding and depth map decoding. Because the enhancement is applied only to image tokens and does not modify camera tokens, camera-parameter regression remains consistent with the backbone VFM. We evaluate DPT-based depth regression after image-token enhancement. As shown in \autoref{tbl_depth_scannet_tat}, depth maps decoded from the enhanced tokens achieve consistent accuracy improvements on both ScanNet and TAT.

\begin{table}[pos=ht]
\centering
\footnotesize
\renewcommand{\arraystretch}{1.02}
\begin{tabularx}{0.49\textwidth}{>{\raggedright\arraybackslash}p{1.65cm}|>{\raggedright\arraybackslash}p{1.45cm}|*{4}{>{\centering\arraybackslash}X}}
\thickhline
\textbf{Dataset} & \textbf{Method} & ABS$\downarrow$ & RMSE$\downarrow$ & $\delta_{1.05}\uparrow$ & $\delta_{1.03}\uparrow$ \\ \thickhline
\multirow{3}{*}{ScanNet\cite{dai2017scannet}}
& VGGT\cite{wang2025vggt} & 0.039 & 0.014 & 0.823 & 0.693 \\
& DA3\cite{depthanything3} & 0.036 & 0.014 & 0.845 & 0.739 \\
& \textbf{$\text{VG}^2$GT} & \textbf{0.032} & \textbf{0.013} & \textbf{0.871} & \textbf{0.788} \\ \thickhline
\multirow{3}{*}{TAT\cite{TAT}}
& VGGT\cite{wang2025vggt} & 0.019 & 0.324 & 0.917 & 0.849 \\
& DA3\cite{depthanything3} & 0.016 & 0.276 & 0.954 & 0.893 \\
& \textbf{$\text{VG}^2$GT} & \textbf{0.013} & \textbf{0.232} & \textbf{0.967} & \textbf{0.927} \\ \thickhline
\end{tabularx}
\caption{Quantitative depth evaluation of coarse-level feature enhancement.}
\label{tbl_depth_scannet_tat}
\end{table}

\subsection{Ablation Study}
To evaluate the contribution of each component in $\text{VG}^2$GT, we conduct ablation studies on continuous stochastic solid volume rendering (3DGS), rendered depth supervision (w/o Depth), coarse-level feature enhancement (w/o Coarse), enhancement loss (w/o Enhance), and the self-splitting Gaussian MLP (w/o Split). Following the same evaluation protocol, we report both NVS quality and geometric reconstruction accuracy for the reconstructed Gaussian scenes.

\begin{table}[pos=ht]
\centering
\scriptsize
\renewcommand{\arraystretch}{1.05}
\begin{tabularx}{0.49\textwidth}{>{\raggedright\arraybackslash}p{1.55cm}|*{3}{>{\centering\arraybackslash}X}|*{4}{>{\centering\arraybackslash}X}}
\thickhline
\multirow{2}{*}{\textbf{Method}} &
\multicolumn{3}{c|}{\textbf{NVS}} &
\multicolumn{4}{c}{\textbf{Depth}} \\
& PSNR & SSIM & LPIPS
& RMSE & ABS & $\delta_{1.25}$ & $\delta_{1.05}$ \\ \thickhline
3DGS & 29.59 & 0.888 & 0.211 & 1.2296 & 0.3005 & 0.7088 & 0.5940 \\
w/o Depth & 28.83 & 0.876 & 0.219 & 0.1014 & 0.0137 & 0.9944 & 0.9855 \\
w/o Coarse & 29.11 & 0.880 & 0.209 & 0.0990 & 0.0130 & 0.9946 & 0.9863 \\
w/o Enhance & 28.93 & 0.878 & 0.219 & 0.0994 & 0.0138 & 0.9946 & 0.9862 \\
w/o Split & 28.19 & 0.865 & {0.219} & 0.1083 & 0.0147 & 0.9935 & 0.9825 \\
\textbf{$\text{VG}^2$GT (full)} & \textbf{30.86} & \textbf{0.898} & \textbf{0.192} & \textbf{0.0786} & \textbf{0.0109} & \textbf{0.9968} & \textbf{0.9904} \\ \thickhline
\end{tabularx}
\caption{Ablation study of $\text{VG}^2$GT. We report averaged NVS and depth metrics on the Replica dataset.}
\label{tbl_ablation}
\end{table}

As shown in \autoref{tbl_ablation}, removing any component degrades both NVS performance and geometric accuracy, indicating that each module contributes to the final reconstruction quality. Ablating continuous stochastic solid volume rendering only slightly reduces NVS performance, but it causes severe geometric degradation. This result confirms its role in improving the geometric accuracy of Gaussian scenes and further shows that better geometry also benefits NVS.

\subsection{Parameter Tuning Experiment}
To validate the splitting number used in fine-level Gaussian scene decoding, we evaluate NVS performance and geometric reconstruction accuracy under different numbers of Gaussian splits. As shown in \autoref{tbl_split}, the default splitting number is 2. Disabling splitting leads to a clear performance drop, whereas increasing the splitting number to 4 provides only marginal NVS gains and slightly degrades rendered-depth accuracy. Because the splitting number is directly proportional to the final number of Gaussian primitives, we use 2 as a practical trade-off among reconstruction quality, computational cost, and storage efficiency.

\begin{table}[pos=ht]
\centering
\scriptsize
\setlength{\tabcolsep}{2.4pt}
\renewcommand{\arraystretch}{1.05}
\begin{tabularx}{0.49\textwidth}{>{\raggedright\arraybackslash}p{1.35cm}|*{3}{>{\centering\arraybackslash}X}|*{4}{>{\centering\arraybackslash}X}}
\thickhline
\multirow{2}{*}{\textbf{Method}} &
\multicolumn{3}{c|}{\textbf{NVS}} &
\multicolumn{4}{c}{\textbf{Depth}} \\
& PSNR & SSIM & LPIPS
& RMSE & ABS & $\delta_{1.25}$ & $\delta_{1.05}$ \\ \thickhline
w/o split & 28.191 & 0.865 & 0.219 & 0.1083 & 0.0147 & 0.9935 & 0.9825 \\
\textbf{split=2} & 30.855 & 0.898 & 0.192 & \textbf{0.0786} & \textbf{0.0109} & \textbf{0.9968} & 0.9904 \\
split=4 & \textbf{30.954} & \textbf{0.899} & \textbf{0.183} & 0.0803 & 0.0111 & 0.9966 & \textbf{0.9906} \\ \thickhline
\end{tabularx}
\caption{Parameter tuning results for the splitting number in fine-level Gaussian scene decoding.}
\label{tbl_split}
\end{table}

\section{Conclusion and further work}
\subsection{Conclusion}
We present $\text{VG}^2$GT, a voxel-Gaussian Splatting visual geometry grounded transformer. By introducing a coarse-to-fine differentiable voxel module, $\text{VG}^2$GT mitigates the overlapping artifacts that commonly arise in pixel-aligned feed-forward Gaussian Splatting methods. At the coarse level, the module enhances image patch tokens and re-decodes geometry-aware features. At the fine level, it splits high-dimensional voxel features to predict geometrically accurate Gaussian scene parameters.
During training, stochastic solid volume rendering supervises scene geometry and improves both geometric accuracy and NVS performance. By using the network as a lightweight add-on to a visual foundation model (VFM), our method substantially reduces training cost while remaining plug-and-play compatible with different VFMs. Experiments across multiple datasets show that $\text{VG}^2$GT reconstructs geometrically accurate Gaussian scenes under different input settings, achieves state-of-the-art RGB and depth NVS performance, and controls the otherwise linear growth of Gaussian primitives as the number of input views increases.
\subsection{Limitation and further work}
The current framework is primarily designed for static scenes. This limitation arises from the explicit representation used by Gaussian Splatting. Directly aggregating observations into a single static voxel-Gaussian representation can therefore introduce blurred geometry, inconsistent Gaussian distributions, or degraded NVS quality.
Extending $\text{VG}^2$GT to dynamic scenes is an important direction for future work. Recent VFMs have shown increasing ability to reason over temporal image sequences \cite{wang2026vggtomega}, while dynamic Gaussian representations can model time-varying motion fields and deformable scene structures \cite{4dgs}. A natural extension is to regress dynamic Gaussian transformation fields directly from temporal VFM features, rather than predicting only static Gaussian parameters. Such a feed-forward formulation could preserve the efficiency of our frozen-backbone design while avoiding the per-scene optimization commonly required by dynamic 3D reconstruction and dynamic NVS methods.

\printcredits

\appendix
\section*{Appendix}

\setcounter{figure}{0}
\setcounter{table}{0}
\setcounter{equation}{0}
\renewcommand{\thefigure}{A\arabic{figure}}
\renewcommand{\thetable}{A\arabic{table}}
\renewcommand{\theequation}{A\arabic{equation}}
\renewcommand{\theHfigure}{appendix.\arabic{figure}}
\renewcommand{\theHtable}{appendix.\arabic{table}}
\renewcommand{\theHequation}{appendix.\arabic{equation}}

\setcounter{section}{0}
\renewcommand{\thesection}{A.\arabic{section}}
\renewcommand{\theHsection}{appendix.\arabic{section}}

\section{Experiment Details}
\hspace{1em}This section provides additional details on model training and evaluation.
\subsection{Model Training}
\hspace{1em}For training initialization, we use DA3-GIANT-1.1 (1.15B) \cite{depthanything3} as the pretrained checkpoint. This model includes a DPT-based pixel-aligned Gaussian head but does not provide a metric-scale head. To recover the scale required by our differentiable multi-scale voxel module, we follow the official GitHub implementation and regress metric depth as $d_{\text{metric}} = \frac{f \cdot d_{\text{net}}}{300}$.

During training, we follow a sampling strategy similar to the official VGGT implementation \cite{wang2025vggt}. To avoid color shifts when training the Gaussian-scene head, we disable random color augmentation.

For loss computation, metric-scale prediction errors in the VFM backbone can destabilize training. We therefore follow AMB3R \cite{wang2025amb3r} and align the rendered depth map $D_{\text{rend}}$ to the ground-truth depth map $D_{\text{gt}}$ with a scale factor $s = \operatorname{wmed}\!\left(D_{\text{gt}} / D_{\text{rend}}\right)$, where $\operatorname{wmed}(\cdot)$ denotes the weighted median. The aligned depth map $\tilde{D}_{\text{rend}} = s D_{\text{rend}}$ is then used to compute the depth loss $\mathcal{L}_\text{depth}$ and the point-map loss $\mathcal{L}_\text{pointmap}$.

\subsection{Model Evaluation}
\hspace{1em}All datasets used in our evaluation, including Replica \cite{replica19arxiv}, DTU \cite{dtu}, Tanks and Temples \cite{TAT}, and ScanNet \cite{dai2017scannet}, consist of sequential frames. We therefore sample input views within a constrained range. Training and evaluation scenes are kept strictly disjoint. For the 3-view evaluation on DTU, we follow the commonly used sparse-view protocol (23, 24, and 33) \cite{huang2025fatesgs,wu2025sparse2dgs} for fair comparison.

During evaluation, all VGGT-based feed-forward methods use the default resolution of $518\times518$ for scene regression and rendering. Because VolSplat, MVSplat, and DepthSplat provide pretrained checkpoints only at $256\times256$, we regress scenes at $256\times256$ and render images at $518\times518$ for fair comparison. For completeness, we also report results when both scene regression and rendering are performed at 256 resolution. As shown in \autoref{tbl_depth_subset_256} and \autoref{tbl_nvs_subset_256}, our method consistently outperforms the baselines in both Gaussian-scene novel view synthesis and depth rendering across different resolution settings.

For the ablation studies, we do not train each ablated variant from scratch, to reduce training cost. Instead, each variant is initialized from the full model and trained for an additional 30K iterations.
For all modules except the self-splitting Gaussian MLP, the network architecture remains unchanged after ablation. Thus, initialization from the full model does not introduce additional architectural or parameter differences. For the self-splitting Gaussian MLP ablation, the final prediction branch has a different output dimension and must be reinitialized. This branch contains only a lightweight two-layer MLP, while the preceding PointMLP-based feature extractor remains unchanged. In this setting, 30K iterations are sufficient for the reinitialized branch to converge.
As shown in \autoref{tbl_ablation}, ablating any individual component degrades performance during continued training, further supporting the effectiveness of each proposed module.

\begin{table*}[pos=ht]
\centering
\footnotesize
\renewcommand{\arraystretch}{0.99}
\begin{tabularx}{\textwidth}{>{\raggedright\arraybackslash}p{1.95cm}|*{4}{>{\centering\arraybackslash}X}|*{4}{>{\centering\arraybackslash}X}|*{4}{>{\centering\arraybackslash}X}}
\thickhline
\multirow{2}{*}{\textbf{Method}} &
\multicolumn{4}{c|}{\textbf{3 Views}} &
\multicolumn{4}{c|}{\textbf{10 Views}} &
\multicolumn{4}{c}{\textbf{20 Views}} \\
& RMSE$\downarrow$ & ABS$\downarrow$ & $\delta_{1.25}\uparrow$ & $\delta_{1.05}\uparrow$
& RMSE$\downarrow$ & ABS$\downarrow$ & $\delta_{1.25}\uparrow$ & $\delta_{1.05}\uparrow$
& RMSE$\downarrow$ & ABS$\downarrow$ & $\delta_{1.25}\uparrow$ & $\delta_{1.05}\uparrow$ \\ \thickhline
\multicolumn{13}{l}{\textit{Replica Dataset\cite{replica19arxiv}}} \\
MVSplat & 0.547 & 0.143 & 0.790 & 0.272 & 0.501 & 0.114 & 0.872 & 0.474 & 0.780 & 0.251 & 0.715 & 0.336 \\
MVSplat\_256 & 0.389 & 0.099 & 0.894 & 0.392 & 0.443 & 0.099 & 0.893 & 0.567 & 0.735 & 0.241 & 0.730 & 0.401 \\
VolSplat & 0.698 & 0.191 & 0.683 & 0.165 & 0.407 & 0.099 & 0.903 & 0.346 & 0.461 & 0.130 & 0.853 & 0.229 \\
VolSplat\_256 & 0.265 & 0.070 & 0.952 & 0.516 & 0.197 & 0.040 & 0.982 & 0.795 & 0.149 & 0.038 & 0.982 & 0.781 \\
DepthSplat & 0.279 & 0.059 & 0.942 & 0.707 & 0.279 & 0.053 & 0.954 & 0.744 & 0.180 & 0.040 & 0.969 & 0.818 \\
DepthSplat\_256 & 0.282 & 0.059 & 0.941 & 0.704 & 0.282 & 0.053 & 0.953 & 0.738 & 0.182 & 0.040 & 0.968 & 0.816 \\ \hline
Ours\_3DGS & 0.092 & 0.011 & 0.995 & 0.940 & 0.109 & 0.014 & 0.996 & 0.980 & 0.100 & 0.022 & 0.980 & 0.966 \\
Ours & 0.069 & 0.007 & 0.997 & 0.990 & 0.078 & 0.011 & 0.997 & 0.990 & 0.043 & 0.006 & 0.999 & 0.994 \\ \thickhline
\multicolumn{13}{l}{\textit{DTU Dataset\cite{dtu}}} \\
MVSplat & 0.069 & 0.663 & 0.163 & 0.037 & 0.092 & 0.093 & 0.905 & 0.430 & 0.140 & 0.129 & 0.836 & 0.398 \\
MVSplat\_256 & 0.067 & 0.619 & 0.194 & 0.044 & 0.083 & 0.087 & 0.920 & 0.414 & 0.115 & 0.103 & 0.887 & 0.455 \\
VolSplat & 0.033 & 0.382 & 0.515 & 0.124 & 0.080 & 0.087 & 0.895 & 0.481 & 0.097 & 0.094 & 0.904 & 0.476 \\
VolSplat\_256 & 0.032 & 0.320 & 0.526 & 0.129 & 0.076 & 0.082 & 0.899 & 0.519 & 0.093 & 0.088 & 0.916 & 0.505 \\
DepthSplat & 0.022 & 0.284 & 0.748 & 0.261 & 0.070 & 0.089 & 0.928 & 0.388 & 0.105 & 0.103 & 0.924 & 0.386 \\
DepthSplat\_256 & 0.022 & 0.227 & 0.749 & 0.260 & 0.070 & 0.089 & 0.928 & 0.388 & 0.106 & 0.104 & 0.924 & 0.386 \\ \hline
Ours\_3DGS & 0.014 & 0.237 & 0.962 & 0.759 & 0.026 & 0.020 & 0.988 & 0.935 & 0.021 & 0.016 & 0.992 & 0.950 \\
Ours & 0.015 & 0.238 & 0.961 & 0.757 & 0.026 & 0.019 & 0.988 & 0.937 & 0.021 & 0.015 & 0.992 & 0.954 \\ \thickhline
\multicolumn{13}{l}{\textit{TAT Dataset\cite{TAT}}} \\
MVSplat & 0.839 & 0.483 & 0.291 & 0.065 & 1.003 & 0.776 & 0.109 & 0.024 & 0.835 & 0.436 & 0.353 & 0.106 \\
MVSplat\_256 & 0.747 & 0.414 & 0.436 & 0.100 & 1.014 & 0.769 & 0.103 & 0.023 & 0.891 & 0.469 & 0.336 & 0.113 \\
VolSplat & 0.673 & 0.355 & 0.556 & 0.164 & 0.348 & 0.321 & 0.498 & 0.138 & 0.418 & 0.200 & 0.693 & 0.234 \\
VolSplat\_256 & 0.557 & 0.300 & 0.712 & 0.326 & 0.294 & 0.281 & 0.600 & 0.208 & 0.367 & 0.170 & 0.764 & 0.344 \\
DepthSplat & 0.488 & 0.295 & 0.723 & 0.351 & 0.431 & 0.291 & 0.655 & 0.229 & 0.413 & 0.179 & 0.733 & 0.354 \\
DepthSplat\_256 & 0.491 & 0.299 & 0.722 & 0.347 & 0.431 & 0.292 & 0.656 & 0.229 & 0.413 & 0.180 & 0.732 & 0.353 \\ \hline
Ours\_3DGS & 0.381 & 0.077 & 0.947 & 0.740 & 0.145 & 0.057 & 0.950 & 0.777 & 0.264 & 0.075 & 0.921 & 0.741 \\
Ours & 0.389 & 0.081 & 0.947 & 0.733 & 0.148 & 0.058 & 0.952 & 0.773 & 0.265 & 0.077 & 0.926 & 0.708 \\ \thickhline
\end{tabularx}
\caption{Quantitative Gaussian-scene depth rendering results on the Replica, DTU, and TAT datasets, including the $256\times256$ variants.}
\label{tbl_depth_subset_256}
\end{table*}

\begin{table*}[pos=ht]
\centering
\footnotesize
\renewcommand{\arraystretch}{1.02}
\begin{tabularx}{0.99\textwidth}{>{\raggedright\arraybackslash}p{1.95cm}|*{3}{>{\centering\arraybackslash}X}|*{3}{>{\centering\arraybackslash}X}|*{3}{>{\centering\arraybackslash}X}}
\thickhline
\multirow{2}{*}{\textbf{Method}} &
\multicolumn{3}{c|}{\textbf{3 Views}} &
\multicolumn{3}{c|}{\textbf{10 Views}} &
\multicolumn{3}{c}{\textbf{20 Views}} \\
& PSNR$\uparrow$ & SSIM$\uparrow$ & LPIPS$\downarrow$
& PSNR$\uparrow$ & SSIM$\uparrow$ & LPIPS$\downarrow$
& PSNR$\uparrow$ & SSIM$\uparrow$ & LPIPS$\downarrow$ \\ \thickhline
\multicolumn{10}{l}{\textit{Replica Dataset\cite{replica19arxiv}}} \\
MVSplat & 25.70 & 0.830 & 0.240 & 24.18 & 0.835 & 0.257 & 24.17 & 0.807 & 0.300 \\
MVSplat\_256 & 27.27 & 0.889 & 0.095 & 23.33 & 0.815 & 0.201 & 23.14 & 0.774 & 0.242 \\
VolSplat & 19.26 & 0.681 & 0.453 & 28.66 & 0.876 & 0.180 & 22.17 & 0.747 & 0.400 \\
VolSplat\_256 & 27.77 & 0.909 & 0.127 & 25.07 & 0.804 & 0.371 & 28.27 & 0.865 & 0.185 \\
DepthSplat & 22.98 & 0.798 & 0.403 & 23.64 & 0.797 & 0.420 & 23.55 & 0.788 & 0.419 \\
DepthSplat\_256 & 23.15 & 0.794 & 0.303 & 23.41 & 0.763 & 0.365 & 23.32 & 0.752 & 0.369 \\ \hline
Ours\_3DGS & 29.87 & 0.876 & 0.184 & 30.08 & 0.883 & 0.208 & 29.54 & 0.876 & 0.202 \\
Ours & 31.03 & 0.898 & 0.169 & 30.87 & 0.898 & 0.191 & 30.41 & 0.894 & 0.184 \\ \thickhline
\multicolumn{10}{l}{\textit{TAT Dataset\cite{TAT}}} \\
MVSplat & 19.83 & 0.628 & 0.359 & 18.06 & 0.545 & 0.441 & 14.62 & 0.412 & 0.584 \\
MVSplat\_256 & 20.92 & 0.650 & 0.252 & 17.71 & 0.499 & 0.372 & 14.42 & 0.338 & 0.552 \\
VolSplat & 16.67 & 0.527 & 0.474 & 17.78 & 0.519 & 0.485 & 16.77 & 0.454 & 0.537 \\
VolSplat\_256 & 20.61 & 0.642 & 0.289 & 19.15 & 0.561 & 0.334 & 16.85 & 0.429 & 0.448 \\
DepthSplat & 16.90 & 0.477 & 0.679 & 15.21 & 0.446 & 0.675 & 14.59 & 0.420 & 0.699 \\
DepthSplat\_256 & 17.12 & 0.407 & 0.640 & 15.19 & 0.355 & 0.653 & 14.59 & 0.328 & 0.671 \\ \hline
Ours\_3DGS & 21.68 & 0.691 & 0.294 & 19.89 & 0.618 & 0.328 & 19.34 & 0.563 & 0.393 \\
Ours & 22.46 & 0.716 & 0.272 & 20.22 & 0.650 & 0.321 & 20.01 & 0.588 & 0.383 \\ \thickhline
\end{tabularx}
\caption{Quantitative NVS results on the Replica and TAT datasets, including the $256\times256$ variants.}
\label{tbl_nvs_subset_256}
\end{table*}

\section{Continuous Stochastic Solids Volume Rendering}

\hspace{1em}We replace vanilla 3DGS rasterization with stochastic solid volume rendering to obtain more accurate scene geometry. Below, we provide the theoretical details.

3DGS and its variants \cite{kerbl3DGS,scaffoldgs} rely on differentiable rasterization to render RGB images and depth maps. As shown in \autoref{eqn_sup_gs_depth_render}, the depth along each ray is computed using the same $\alpha$-blending formulation as RGB rendering. Although this design preserves differentiability for Gaussian primitives along the entire ray, it also blends foreground and background depths, which can produce floating-depth artifacts and degrade geometric reconstruction accuracy \cite{Huang2DGS2024}.

\begin{equation}
\label{eqn_sup_gs_depth_render}
\begin{aligned}
D_{\text{expected}} 
= \frac{\displaystyle\sum_{i=1}^{N} d_i \cdot \alpha_i \cdot T_i}
       {1 - T_N}
\end{aligned}
\end{equation}

2DGS and its variants \cite{Huang2DGS2024,huang2025fatesgs} use per-pixel median depth as the final rendered depth. This depth corresponds to the Gaussian primitive for which the accumulated transmittance $T_i = \prod_{j=1}^{i-1}(1 - \alpha_j)$ first falls below 0.5. This strategy improves geometric reconstruction accuracy. However, fixed median-depth selection can limit further gains in geometric precision. Boundary regions of Gaussian primitives can also introduce discontinuities in the rendered depth map.
Moreover, median-depth rendering affects only a single median Gaussian primitive during backpropagation, which is unfavorable for optimizing generalizable foundation models.

To address this issue, we replace vanilla rasterization with continuous stochastic solids volume rendering \cite{zhang2026gggs}. Following \cite{2023Objectsasvolumes}, for a stochastic solid with vacancy $v$, the attenuation coefficient $\sigma$ is defined as in \autoref{eqn_sup_attenuation_coefficient}.
\begin{equation}
\label{eqn_sup_attenuation_coefficient}
\begin{aligned}
\sigma(\mathbf{x}, \omega)=\left| \omega \cdot \nabla \log\left(v(\mathbf{x})\right) \right|=\frac{\left| \omega \cdot \nabla v(\mathbf{x}) \right|}{v(\mathbf{x})}
\end{aligned}
\end{equation}
Here, $\mathbf{x}$ denotes the 3D spatial position and $\omega$ denotes the ray direction. The volume rendering formulation for stochastic solids is shown in \autoref{eqn_volume_render}.

Following the derivation in \cite{zhang2026gggs}, we treat each Gaussian primitive as a continuous stochastic solid. We interpret the per-pixel opacity $\alpha$ as the maximum value of the Gaussian primitive along the pixel ray, as shown in \autoref{eqn_gggs_color}. This leads to the vacancy formulation in \autoref{eqn_vacancy}, which establishes an equivalent conversion from a 3D Gaussian distribution to a Gaussian primitive and the corresponding volume rendering formulation.

\bibliographystyle{cas-model2-names}

\bibliography{cas-refs}

@inproceedings{wang2025vggt,
  title={VGGT: Visual Geometry Grounded Transformer},
  author={Wang, Jianyuan and Chen, Minghao and Karaev, Nikita and Vedaldi, Andrea and Rupprecht, Christian and Novotny, David},
  booktitle={Proceedings of the IEEE/CVF Conference on Computer Vision and Pattern Recognition},
  year={2025}
}

@article{depthanything3,
  title={Depth Anything 3: Recovering the visual space from any views},
  author={Haotong Lin and Sili Chen and Jun Hao Liew and Donny Y. Chen and Zhenyu Li and Guang Shi and Jiashi Feng and Bingyi Kang},
  journal={arXiv preprint arXiv:2511.10647},
  year={2025}
}

@article{DPT,
	author    = {Ren\'{e} Ranftl and Alexey Bochkovskiy and Vladlen Koltun},
	title     = {Vision Transformers for Dense Prediction},
	journal   = {ArXiv preprint},
	year      = {2021},
}

@misc{oquab2023dinov2,
  title={DINOv2: Learning Robust Visual Features without Supervision},
  author={Oquab, Maxime and Darcet, Timothée and Moutakanni, Theo and Vo, Huy V. and Szafraniec, Marc and Khalidov, Vasil and Fernandez, Pierre and Haziza, Daniel and Massa, Francisco and El-Nouby, Alaaeldin and Howes, Russell and Huang, Po-Yao and Xu, Hu and Sharma, Vasu and Li, Shang-Wen and Galuba, Wojciech and Rabbat, Mike and Assran, Mido and Ballas, Nicolas and Synnaeve, Gabriel and Misra, Ishan and Jegou, Herve and Mairal, Julien and Labatut, Patrick and Joulin, Armand and Bojanowski, Piotr},
  journal={arXiv:2304.07193},
  year={2023}
}

@inproceedings{zhu2025voxelsplat,
  title={Voxelsplat: Dynamic gaussian splatting as an effective loss for occupancy and flow prediction},
  author={Zhu, Ziyue and Wang, Shenlong and Xie, Jin and Liu, Jiang-jiang and Wang, Jingdong and Yang, Jian},
  booktitle={Proceedings of the Computer Vision and Pattern Recognition Conference},
  pages={6761--6771},
  year={2025}
}

@article{wang2025amb3r,
  title={AMB3R: Accurate Feed-forward Metric-scale 3D Reconstruction with Backend},
  author={Wang, Hengyi and Agapito, Lourdes},
  journal={arXiv preprint arXiv:2511.20343},
  year={2025}
}

@inproceedings{wu2024ptv3,
    title={Point Transformer V3: Simpler, Faster, Stronger},
    author={Wu, Xiaoyang and Jiang, Li and Wang, Peng-Shuai and Liu, Zhijian and Liu, Xihui and Qiao, Yu and Ouyang, Wanli and He, Tong and Zhao, Hengshuang},
    booktitle={CVPR},
    year={2024}
}

@article{pointmlp,
    title={Rethinking network design and local geometry in point cloud: A simple residual MLP framework},
    author={Ma, Xu and Qin, Can and You, Haoxuan and Ran, Haoxi and Fu, Yun},
    journal={arXiv preprint arXiv:2202.07123},
    year={2022}
}

@article{2023Objectsasvolumes,
  title={Objects as volumes: A stochastic geometry view of opaque solids},
  author={ Miller, Bailey  and  Chen, Hanyu  and  Lai, Alice  and  Gkioulekas, Ioannis },
  journal={IEEE},
  year={2023},
}

@article{zhang2026gggs,
  title={Geometry-Grounded Gaussian Splatting},
  author={Zhang, Baowen and Jiang, Chenxing and Li, Heng and Shen, Shaojie and Tan, Ping},
  journal={arXiv preprint arXiv:2601.17835},
  year={2026}
}

@Article{kerbl3DGS,
      author       = {Kerbl, Bernhard and Kopanas, Georgios and Leimk{\"u}hler, Thomas and Drettakis, George},
      title        = {3D Gaussian Splatting for Real-Time Radiance Field Rendering},
      journal      = {ACM Transactions on Graphics},
      number       = {4},
      volume       = {42},
      month        = {July},
      year         = {2023},
      url          = {https://repo-sam.inria.fr/fungraph/3d-gaussian-splatting/}
}

@inproceedings{Huang2DGS2024,
    title={2D Gaussian Splatting for Geometrically Accurate Radiance Fields},
    author={Huang, Binbin and Yu, Zehao and Chen, Anpei and Geiger, Andreas and Gao, Shenghua},
    publisher = {Association for Computing Machinery},
    booktitle = {SIGGRAPH 2024 Conference Papers},
    year      = {2024},
    doi       = {10.1145/3641519.3657428}
}

@inproceedings{huang2025fatesgs,
    title={FatesGS: Fast and Accurate Sparse-View Surface Reconstruction Using Gaussian Splatting with Depth-Feature Consistency},
    author={Han Huang and Yulun Wu and Chao Deng and Ge Gao and Ming Gu and Yu-Shen Liu},
    booktitle={Proceedings of the AAAI Conference on Artificial Intelligence},
    year={2025}
}

@article{wang2004ssim,
  title={Image quality assessment: from error visibility to structural similarity},
  author={Wang, Zhou and Bovik, Alan C and Sheikh, Hamid R and Simoncelli, Eero P},
  journal={IEEE transactions on image processing},
  volume={13},
  number={4},
  pages={600--612},
  year={2004},
  publisher={IEEE}
}

@inproceedings{dust3r_cvpr24,
      title={DUSt3R: Geometric 3D Vision Made Easy}, 
      author={Shuzhe Wang and Vincent Leroy and Yohann Cabon and Boris Chidlovskii and Jerome Revaud},
      booktitle = {CVPR},
      year = {2024}
}

@article{sift,
  title={Scale invariant feature transform},
  author={Lindeberg, Tony},
  year={2012}
}

@inproceedings{detone2018superpoint,
  title={Superpoint: Self-supervised interest point detection and description},
  author={DeTone, Daniel and Malisiewicz, Tomasz and Rabinovich, Andrew},
  booktitle={Proceedings of the IEEE conference on computer vision and pattern recognition workshops},
  pages={224--236},
  year={2018}
}

@inproceedings{sarlin2020superglue,
  title={Superglue: Learning feature matching with graph neural networks},
  author={Sarlin, Paul-Edouard and DeTone, Daniel and Malisiewicz, Tomasz and Rabinovich, Andrew},
  booktitle={Proceedings of the IEEE/CVF conference on computer vision and pattern recognition},
  pages={4938--4947},
  year={2020}
}

@inproceedings{plane-sweeping,
  title={Real-time plane-sweeping stereo with multiple sweeping directions},
  author={Gallup, David and Frahm, Jan-Michael and Mordohai, Philippos and Yang, Qingxiong and Pollefeys, Marc},
  booktitle={2007 IEEE conference on computer vision and pattern recognition},
  pages={1--8},
  year={2007},
  organization={IEEE}
}

@inproceedings{colmapmvs,
    author={Sch\"{o}nberger, Johannes Lutz and Zheng, Enliang and Pollefeys, Marc and Frahm, Jan-Michael},
    title={Pixelwise View Selection for Unstructured Multi-View Stereo},
    booktitle={European Conference on Computer Vision (ECCV)},
    year={2016},
}

@article{yao2018mvsnet,
  title={MVSNet: Depth Inference for Unstructured Multi-view Stereo},
  author={Yao, Yao and Luo, Zixin and Li, Shiwei and Fang, Tian and Quan, Long},
  journal={European Conference on Computer Vision (ECCV)},
  year={2018}
}

@inproceedings{mildenhall2020nerf,
  title={NeRF: Representing Scenes as Neural Radiance Fields for View Synthesis},
  author={Ben Mildenhall and Pratul P. Srinivasan and Matthew Tancik and Jonathan T. Barron and Ravi Ramamoorthi and Ren Ng},
  year={2020},
  booktitle={ECCV},
}

@article{zybapin2025robust,
  title={Robust Geometric Reconstruction of RGB-D Data Based on Gaussian Splatting},
  author={Zhao, Yibin and Yi, Jianjun and Pan, Yihan and Chen, Liwei},
  journal={Applied Intelligence},
  volume={55},
  number={17},
  pages={1118},
  year={2025},
  publisher={Springer}
}

@inproceedings{keetha2026mapanything,
  title={{MapAnything}: Universal Feed-Forward Metric {3D} Reconstruction},
  author={Nikhil Keetha and Norman M\"{u}ller and Johannes Sch\"{o}nberger and Lorenzo Porzi and Yuchen Zhang and Tobias Fischer and Arno Knapitsch and Duncan Zauss and Ethan Weber and Nelson Antunes and Jonathon Luiten and Manuel Lopez-Antequera and Samuel Rota Bul\`{o} and Christian Richardt and Deva Ramanan and Sebastian Scherer and Peter Kontschieder},
  booktitle={International Conference on 3D Vision (3DV)},
  year={2026},
  organization={IEEE}
}

@article{wang2025pi3,
  title={{$\pi^3$: Permutation-Equivariant Visual Geometry Learning}},
  author={Wang, Yifan and Zhou, Jianjun and Zhu, Haoyi and Chang, Wenzheng and Zhou, Yang and Li, Zizun and Chen, Junyi and Pang, Jiangmiao and Shen, Chunhua and He, Tong},
  journal={arXiv preprint arXiv:2507.13347},
  year={2025}
}

@inproceedings{barron2022mipnerf360,
  title={Mip-nerf 360: Unbounded anti-aliased neural radiance fields},
  author={Barron, Jonathan T and Mildenhall, Ben and Verbin, Dor and Srinivasan, Pratul P and Hedman, Peter},
  booktitle={Proceedings of the IEEE/CVF conference on computer vision and pattern recognition},
  pages={5470--5479},
  year={2022}
}

@inproceedings{scaffoldgs,
  title={Scaffold-gs: Structured 3d gaussians for view-adaptive rendering},
  author={Lu, Tao and Yu, Mulin and Xu, Linning and Xiangli, Yuanbo and Wang, Limin and Lin, Dahua and Dai, Bo},
  booktitle={Proceedings of the IEEE/CVF Conference on Computer Vision and Pattern Recognition},
  pages={20654--20664},
  year={2024}
}

@article{ren2025fastgs,
  title={FastGS: Training 3D Gaussian Splatting in 100 Seconds},
  author={Ren, Shiwei and Wen, Tianci and Fang, Yongchun and Lu, Biao},
  journal={arXiv preprint arXiv:2511.04283},
  year={2025}
}

@inproceedings{colmapsfm,
    author={Sch\"{o}nberger, Johannes Lutz and Frahm, Jan-Michael},
    title={Structure-from-Motion Revisited},
    booktitle={Conference on Computer Vision and Pattern Recognition (CVPR)},
    year={2016},
}

@inproceedings{glomap,
  title={Global structure-from-motion revisited},
  author={Pan, Linfei and Bar{\'a}th, D{\'a}niel and Pollefeys, Marc and Sch{\"o}nberger, Johannes L},
  booktitle={European Conference on Computer Vision},
  pages={58--77},
  year={2024},
  organization={Springer}
}

@inproceedings{wang2024vggsfm,
  title={Vggsfm: Visual geometry grounded deep structure from motion},
  author={Wang, Jianyuan and Karaev, Nikita and Rupprecht, Christian and Novotny, David},
  booktitle={Proceedings of the IEEE/CVF conference on computer vision and pattern recognition},
  pages={21686--21697},
  year={2024}
}

@inproceedings{keetha2024splatam,
  title={Splatam: Splat track \& map 3d gaussians for dense rgb-d slam},
  author={Keetha, Nikhil and Karhade, Jay and Jatavallabhula, Krishna Murthy and Yang, Gengshan and Scherer, Sebastian and Ramanan, Deva and Luiten, Jonathon},
  booktitle={Proceedings of the IEEE/CVF conference on computer vision and pattern recognition},
  pages={21357--21366},
  year={2024}
}

@inproceedings{gsicpslam,
  title={Rgbd gs-icp slam},
  author={Ha, Seongbo and Yeon, Jiung and Yu, Hyeonwoo},
  booktitle={European conference on computer vision},
  pages={180--197},
  year={2024},
  organization={Springer}
}

@inproceedings{meng2021gnerf,
  title={Gnerf: Gan-based neural radiance field without posed camera},
  author={Meng, Quan and Chen, Anpei and Luo, Haimin and Wu, Minye and Su, Hao and Xu, Lan and He, Xuming and Yu, Jingyi},
  booktitle={Proceedings of the IEEE/CVF International Conference on Computer Vision},
  pages={6351--6361},
  year={2021}
}

@article{wang2021nerf--,
  title={NeRF--: Neural radiance fields without known camera parameters},
  author={Wang, Zirui and Wu, Shangzhe and Xie, Weidi and Chen, Min and Prisacariu, Victor Adrian},
  year={2021}
}

@article{xu2024freesplatter,
  title={FreeSplatter: Pose-free Gaussian Splatting for Sparse-view 3D Reconstruction},
  author={Xu, Jiale and Gao, Shenghua and Shan, Ying},
  journal={arXiv preprint arXiv:2412.09573},
  year={2024}
}

@article{li2026tokensplat,
  title={TokenSplat: Token-aligned 3D Gaussian Splatting for Feed-forward Pose-free Reconstruction},
  author={Li, Yihui and Lv, Chengxin and Tang, Zichen and Yang, Hongyu and Huang, Di},
  journal={arXiv preprint arXiv:2603.00697},
  year={2026}
}

@article{jiang2025anysplat,
  title={Anysplat: Feed-forward 3d gaussian splatting from unconstrained views},
  author={Jiang, Lihan and Mao, Yucheng and Xu, Linning and Lu, Tao and Ren, Kerui and Jin, Yichen and Xu, Xudong and Yu, Mulin and Pang, Jiangmiao and Zhao, Feng and others},
  journal={ACM Transactions on Graphics (TOG)},
  volume={44},
  number={6},
  pages={1--16},
  year={2025},
  publisher={ACM New York, NY, USA}
}

@misc{ye2025yonosplatneedmodelfeedforward,
  title={YoNoSplat: You Only Need One Model for Feedforward 3D Gaussian Splatting}, 
  author={Botao Ye and Boqi Chen and Haofei Xu and Daniel Barath and Marc Pollefeys},
  year={2025},
  eprint={2511.07321},
  archivePrefix={arXiv},
  primaryClass={cs.CV} 
}

@inproceedings{xu2024depthsplat,
      title   = {DepthSplat: Connecting Gaussian Splatting and Depth},
      author  = {Xu, Haofei and Peng, Songyou and Wang, Fangjinhua and Blum, Hermann and Barath, Daniel and Geiger, Andreas and Pollefeys, Marc},
      booktitle={CVPR},
      year={2025}
    }

@article{chen2024mvsplat,
    title   = {MVSplat: Efficient 3D Gaussian Splatting from Sparse Multi-View Images},
    author  = {Chen, Yuedong and Xu, Haofei and Zheng, Chuanxia and Zhuang, Bohan and Pollefeys, Marc and Geiger, Andreas and Cham, Tat-Jen and Cai, Jianfei},
    journal = {arXiv preprint arXiv:2403.14627},
    year    = {2024},
}

@article{wang2025volsplat,
  title={VolSplat: Rethinking Feed-Forward 3D Gaussian Splatting with Voxel-Aligned Prediction},
  author={Wang, Weijie and Chen, Yeqing and Zhang, Zeyu and Liu, Hengyu and Wang, Haoxiao and Feng, Zhiyuan and Qin, Wenkang and Chen, Feng and Zhu, Zheng and Chen, Donny Y. and Zhuang, Bohan},
  journal={arXiv preprint arXiv:2509.19297},
  year={2025}
}

@article{li2025iggt,
  title={IGGT: Instance-Grounded Geometry Transformer for Semantic 3D Reconstruction},
  author={Li, Hao and Zou, Zhengyu and Liu, Fangfu and Zhang, Xuanyang and Hong, Fangzhou and Cao, Yukang and Lan, Yushi and Zhang, Manyuan and Yu, Gang and Zhang, Dingwen and others},
  journal={arXiv preprint arXiv:2510.22706},
  year={2025}
}

@inproceedings{kittidepth,
  author = {Jonas Uhrig and Nick Schneider and Lukas Schneider and Uwe Franke and Thomas Brox and Andreas Geiger},
  title = {Sparsity Invariant CNNs},
  booktitle = {International Conference on 3D Vision (3DV)},
  year = {2017}
}

@article{kittiraw,
  author = {Andreas Geiger and Philip Lenz and Christoph Stiller and Raquel Urtasun},
  title = {Vision meets Robotics: The KITTI Dataset},
  journal = {International Journal of Robotics Research (IJRR)},
  year = {2013}
}

@inproceedings{dai2017scannet,
    title={ScanNet: Richly-annotated 3D Reconstructions of Indoor Scenes},
    author={Dai, Angela and Chang, Angel X. and Savva, Manolis and Halber, Maciej and Funkhouser, Thomas and Nie{\ss}ner, Matthias},
    booktitle = {Proc. Computer Vision and Pattern Recognition (CVPR), IEEE},
    year = {2017}
}

@article{replica19arxiv,
  title =   {The {R}eplica Dataset: A Digital Replica of Indoor Spaces},
  author =  {Julian Straub and Thomas Whelan and Lingni Ma and Yufan Chen and Erik Wijmans and Simon Green and Jakob J. Engel and Raul Mur-Artal and Carl Ren and Shobhit Verma and Anton Clarkson and Mingfei Yan and Brian Budge and Yajie Yan and Xiaqing Pan and June Yon and Yuyang Zou and Kimberly Leon and Nigel Carter and Jesus Briales and  Tyler Gillingham and  Elias Mueggler and Luis Pesqueira and Manolis Savva and Dhruv Batra and Hauke M. Strasdat and Renzo De Nardi and Michael Goesele and Steven Lovegrove and Richard Newcombe },
  journal = {arXiv preprint arXiv:1906.05797},
  year =    {2019}
}

@article{dtu,
  title={Large-Scale Data for Multiple-View Stereopsis},
  author={Aanaes and HenrikJensen and Rasmus RamsbolVogiatzis and GeorgeTola and EnginDahl and Anders Bjorholm},
  journal={International Journal of Computer Vision},
  volume={120},
  number={2},
  year={2016},
}

@article{TAT,
    author    = {Arno Knapitsch and Jaesik Park and Qian-Yi Zhou and Vladlen Koltun},
    title     = {Tanks and Temples: Benchmarking Large-Scale Scene Reconstruction},
    journal   = {ACM Transactions on Graphics},
    volume    = {36},
    number    = {4},
    year      = {2017},
}

@article{SSIM,
  title={Image quality assessment: from error visibility to structural similarity},
  author={ Wang, Zhou  and  Bovik, Alan Conrad  and  Sheikh, Hamid Rahim  and  Simoncelli, Eero P. },
  journal={IEEE Trans Image Process},
  volume={13},
  number={4},
  year={2004},
}

@article{LPIPS,
  title={The Unreasonable Effectiveness of Deep Features as a Perceptual Metric},
  author={ Zhang, Richard  and  Isola, Phillip  and  Efros, Alexei A  and  Shechtman, Eli  and  Wang, Oliver },
  journal={IEEE},
  year={2018},
}

@inproceedings{dao2023flashattention2,
  title={Flash{A}ttention-2: Faster Attention with Better Parallelism and Work Partitioning},
  author={Dao, Tri},
  booktitle={International Conference on Learning Representations (ICLR)},
  year={2024}
}

@article{wu2025sparse2dgs,
  title={Sparse2DGS: Geometry-Prioritized Gaussian Splatting for Surface Reconstruction from Sparse Views},
  author={Wu, Jiang and Li, Rui and Zhu, Yu and Guo, Rong and Sun, Jinqiu and Zhang, Yanning},
  journal={arXiv preprint arXiv:2504.20378},
  year={2025}
}

@misc{wang2026vggtomega,
      title={VGGT-$\Omega$}, 
      author={Jianyuan Wang and Minghao Chen and Shangzhan Zhang and Nikita Karaev and Johannes Schönberger and Patrick Labatut and Piotr Bojanowski and David Novotny and Andrea Vedaldi and Christian Rupprecht},
      year={2026},
      eprint={2605.15195},
      archivePrefix={arXiv},
      primaryClass={cs.CV},
      url={https://arxiv.org/abs/2605.15195}, 
}

@InProceedings{4dgs,
    author    = {Wu, Guanjun and Yi, Taoran and Fang, Jiemin and Xie, Lingxi and Zhang, Xiaopeng and Wei, Wei and Liu, Wenyu and Tian, Qi and Wang, Xinggang},
    title     = {4D Gaussian Splatting for Real-Time Dynamic Scene Rendering},
    booktitle = {Proceedings of the IEEE/CVF Conference on Computer Vision and Pattern Recognition (CVPR)},
    month     = {June},
    year      = {2024},
    pages     = {20310-20320}
}


\bio{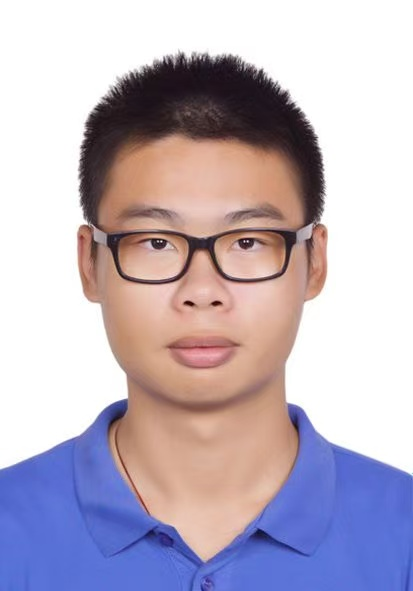}
\textbf{Yibin Zhao} is currently a Ph.D. student at East China University of Science and Technology at Shanghai. 
He received his bachelor degree from East China University of Science and Technology in 2023. 
His research interests include computer vision and 3D reconstruction. 
Now he mainly works on the area of NVS and dense 3D reconstruction. 
\endbio
\vspace{2.5em}

\bio{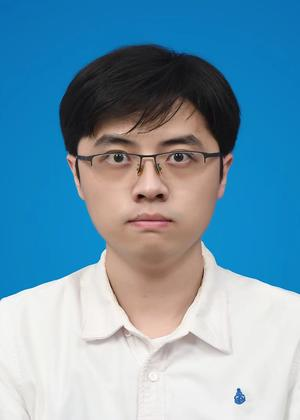}
\textbf{Yihan Pan} is currently a Ph.D. student at East China University of Science and Technology at Shanghai. 
He received his bachelor degree from East China University of Science and Technology in 2021. 
His research interests include computer vision and 3D reconstruction. 
Now he mainly works on the area of point cloud registration. 
\endbio
\vspace{2.5em}

\bio{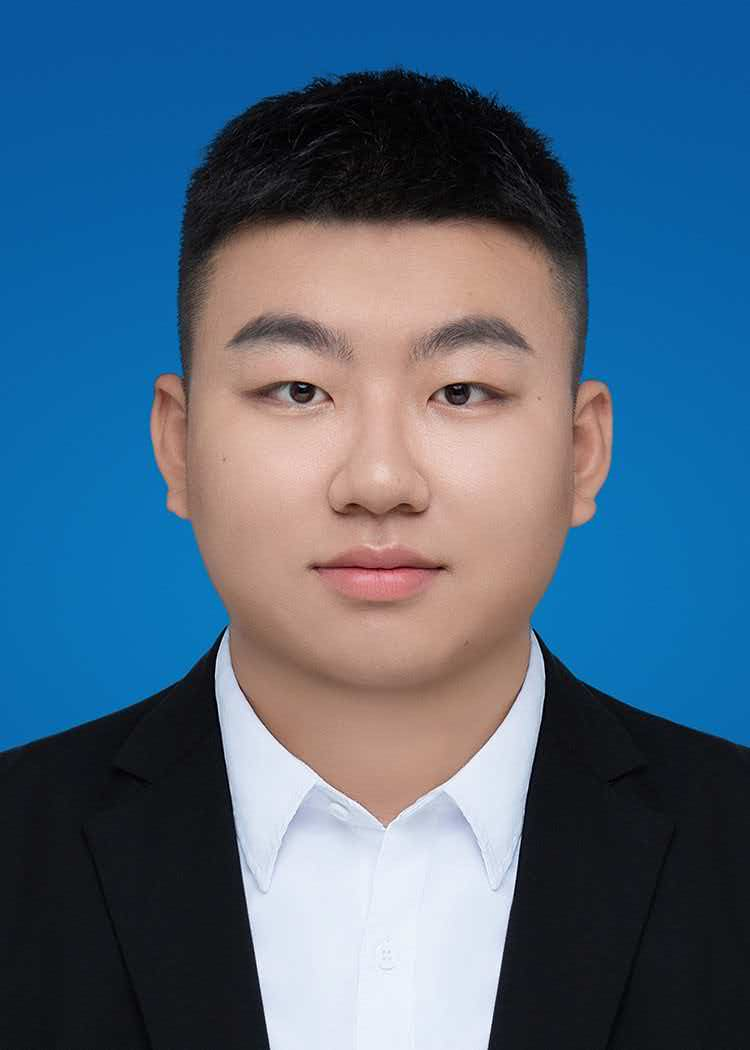}
\textbf{Jun Nan} is currently a Ph.D. student at East China University of Science and Technology at Shanghai. 
He received his bachelor degree from East China University of Science and Technology in 2024. 
His research interests include robotics. 
Now he mainly works on the area of embodied intelligence. 
\endbio
\vspace{3.5em}

\bio{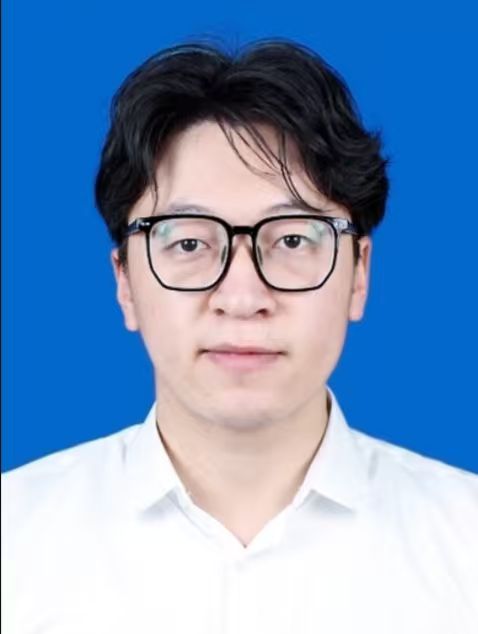}
\textbf{Wenli Yang} is currently a Master's student at East China University of Science and Technology at Shanghai. 
He received his bachelor degree from East China University of Science and Technology in 2026. 
His research interests include 3D reconstruction and semantic segmentation. 
Now he mainly works on the area of 3DCV and Stereo Vision. 
\endbio
\vspace{6em}

\bio{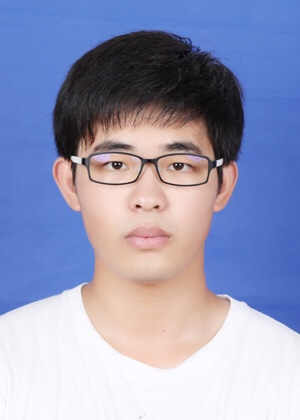}
\textbf{Liwei Chen} is currently a Ph.D. student at East China University of Science and Technology at Shanghai. 
He received his bachelor degree from East China University of Science and Technology in 2019. 
His research interests include semantic segmentation and deep learning. 
Now he mainly works on the area of GAN and target identification. 
\endbio
\vspace{2.5em}

\bio{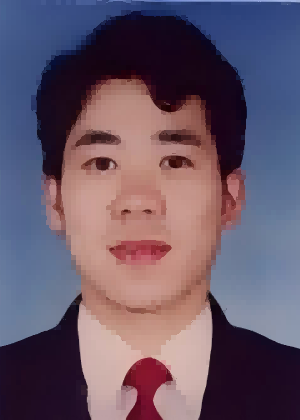}
\textbf{Jianjun Yi} is a professor with the School of Mechanical and Power Engineering and Computer Science, East China University of Science and Technology, Shanghai, China. His main research interests include intelligent control System, mechatronics, smart sensing and intelligent recognition, artificial intelligence, embedded Control System and so on. 
\endbio
\vspace{2.5em}

\end{document}